\documentclass[10pt]{article}
\usepackage[T1]{fontenc}
\usepackage[utf8]{inputenc}
\usepackage[margin=1in]{geometry}
\usepackage{amsmath,mathtools}
\usepackage{XCharter}
\usepackage[xcharter]{newtxmath}
\usepackage[scaled=0.9]{helvet}
\usepackage{microtype}
\usepackage{booktabs}
\usepackage{graphicx}
\usepackage[dvipsnames]{xcolor}
\usepackage{float}
\usepackage{algorithm}
\usepackage{algpseudocode}
\usepackage[round]{natbib}
\usepackage[skins]{tcolorbox}
\usepackage{titlesec}
\usepackage[font=small]{caption}
\definecolor{linkblue}{HTML}{1F4FBF}
\definecolor{boxgray}{HTML}{F3F3F4}
\usepackage[colorlinks=true,linkcolor=linkblue,citecolor=linkblue,urlcolor=linkblue]{hyperref}
\usepackage{url}

\titleformat{\section}{\sffamily\bfseries\Large}{\thesection}{0.8em}{}
\titleformat{\subsection}{\sffamily\bfseries\large}{\thesubsection}{0.8em}{}
\titleformat{\paragraph}[runin]{\sffamily\bfseries}{}{0pt}{}[.]
\titlespacing*{\paragraph}{0pt}{1.0ex plus .5ex minus .2ex}{0.6em}
\newcommand{\tightdisplay}{%
  \setlength{\abovedisplayskip}{3pt plus1pt minus1pt}%
  \setlength{\belowdisplayskip}{3pt plus1pt minus1pt}%
  \setlength{\abovedisplayshortskip}{0pt}%
  \setlength{\belowdisplayshortskip}{2pt plus1pt minus1pt}}
\newcommand{\avgk}{avg@$K$}
\newcommand{\passk}{pass@$K$}
\newcommand{\tsopd}{TS-OPD}
\newcommand{\tsro}{TS-OPD (routing only)}
\newcommand{\AC}{A$\cup$C}
\newcommand{\Gt}{G_{\mathrm{transfer}}}
\newcommand{\qt}{\operatorname{qt}}
\newcommand{\tA}{\mathsf{A}}
\newcommand{\tC}{\mathsf{C}}
\newcommand{\tP}{\mathsf{P}}
\newcommand{\tN}{\mathsf{N}}

\makeatletter\g@addto@macro\normalsize{\tightdisplay}\makeatother
\begin{document}
\tightdisplay

\thispagestyle{plain}
\begin{tcolorbox}[enhanced, colback=boxgray, colframe=boxgray, boxrule=0pt,
  arc=3.5mm, left=7mm, right=7mm, top=6mm, bottom=5mm]
{\sffamily\bfseries\LARGE Transfer-Stratified On-Policy Distillation\\[2pt]
for RL-Improved Reasoning Teachers\par}
\vspace{3mm}
{\sffamily\bfseries\small Xiaoyu Chen$^{1,2,*}$, Bo Shao$^{2}$, Tiangang Zhu$^{2}$, Bintao Wu$^{3}$,
Linjun Shou$^{2}$, Fengge Wu$^{1}$, Feng Sun$^{2}$, Wenbiao Ding$^{2}$\par}
\vspace{2.5mm}
{\small $^{1}$Institute of Software, Chinese Academy of Sciences,
$^{2}$Microsoft, $^{3}$Northwestern Polytechnical University\par
$^{*}$Work done during an internship at Microsoft\par}
\vspace{4mm}
{\small
Reinforcement learning can substantially improve a reasoning teacher, but it is
unclear which of those improvements survive when the teacher supervises a
smaller on-policy student. We study this question in mathematical reasoning by
comparing teacher lineages before and after GRPO, multiple student scales,
direct GRPO, and several on-policy distillation objectives. The central finding
is that transfer is structured rather than scalar: teacher strength alone does
not make dense distillation competitive, while an RL-improved teacher creates
useful but metric-dependent student gains. This motivates Transfer-Stratified
On-Policy Distillation (\tsopd), which screens training problems by the joint
sampled success of the student and teacher, routes acquisition problems to
gated forward KL, routes consolidation problems to gated reverse KL, and adds an
entropy brake to protect sampled coverage. Across the main comparison,
\tsopd{} is the strongest student objective for macro average correctness with
the GRPO-improved teacher, while pass@$K$ remains more mixed. Ablations show
that the gains come from routing and token gating rather than skipping problems. These
results support a transfer-aware view of OPD: stronger teachers help when the
supervision direction and token budget match the student's observed ability,
not merely because the teacher endpoint is stronger.\par}
\vspace{4mm}
{\footnotesize\sffamily\textbf{Correspondence:}}~{\small Wenbiao Ding
(\href{mailto:wenbiaoding@microsoft.com}{wenbiaoding@microsoft.com})}%
\hfill\includegraphics[width=4.68cm]{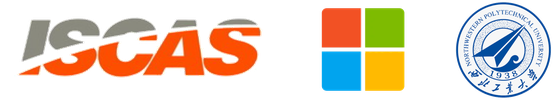}
\end{tcolorbox}
\vspace{4mm}

\section{Introduction}

On-policy distillation (OPD) is the dominant paradigm for transferring
reasoning capabilities from large models to smaller ones
\citep{agarwal2024opd,gu2024minillm}. It trains students on their own generated
states under token-level supervision from a frozen teacher, reducing the
training--inference distribution mismatch. A teacher improved through
reinforcement learning with verifiable rewards, such as GRPO
\citep{shao2024deepseekmath,guo2025deepseekr1}, should therefore provide stronger
supervision. Prior work indeed shows that a GRPO-trained teacher produces a
stronger student than the original teacher \citep{sang2026beyond}. Yet our
initial observations on MATH \citep{hendrycks2021math} show that student gains
can fall short of the teacher's GRPO improvement. We call this shortfall the
\textbf{transfer gap} and ask: \textbf{which teacher improvements transfer to
the student, which do not, and how can more of them be made to transfer?}

\begin{figure}[t]
\centering
\includegraphics[width=\textwidth]{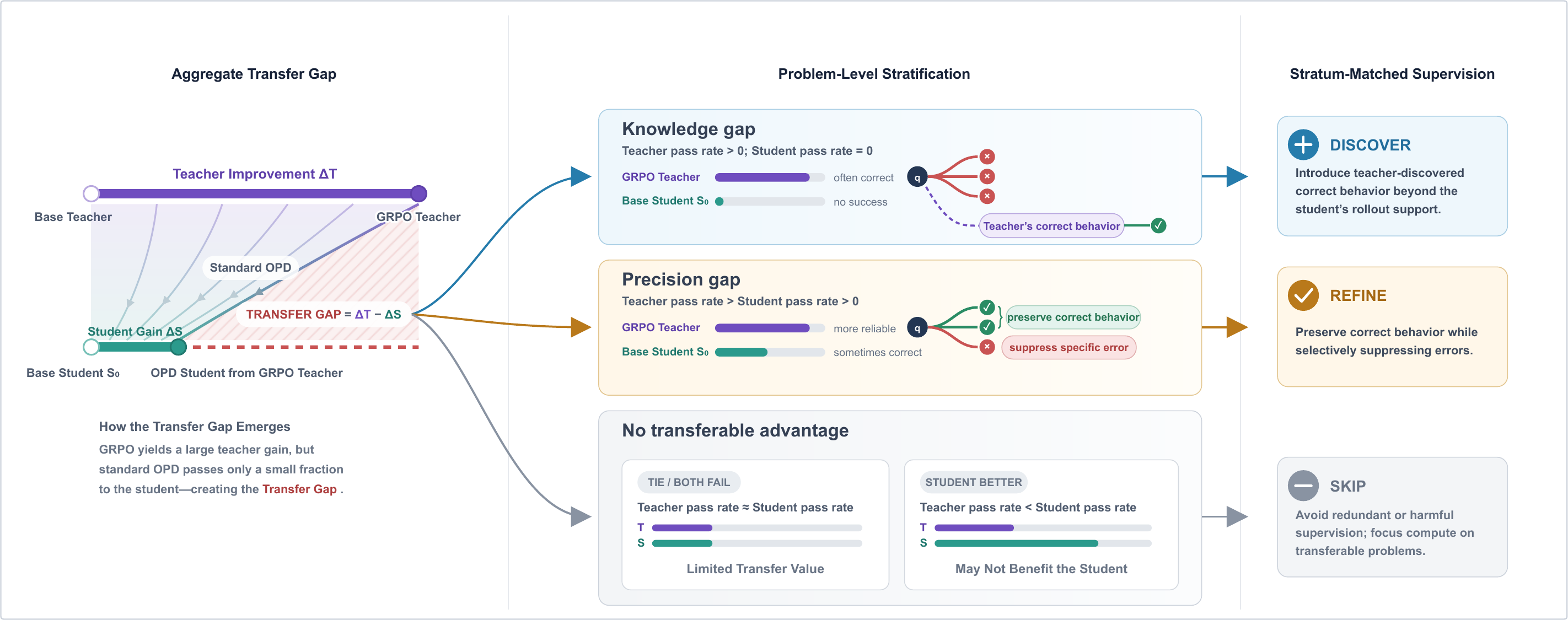}
\caption{\textbf{Overview of the transfer gap and its problem-level
decomposition.} \textbf{Left:} GRPO yields a teacher improvement $\Delta T$,
but standard OPD transfers only part of it into a student gain $\Delta S$,
leaving a transfer gap of $\Delta T-\Delta S$. \textbf{Middle:} Joint
teacher--student pass rates identify knowledge gaps (teacher $>0$, student
$=0$), precision gaps (teacher $>$ student $>0$), and problems with no
transferable teacher advantage (teacher $\leq$ student). \textbf{Right:}
TS-OPD routes these three strata to matched actions---discover, refine, and
skip, respectively.}
\label{fig:intro}
\end{figure}

Our problem-level analysis on MATH shows that this gap is not uniform across
training problems (Figure~\ref{fig:intro}). Whether a teacher improvement can
transfer depends on how the GRPO-trained teacher (hereafter, the \emph{GRPO
teacher}) and student perform on each problem. Where the teacher outperforms
the student, two distinct learning needs arise. If the student never solves
the problem, it faces a \textbf{knowledge gap}: it needs guidance toward
correct choices it has never explored, yet standard OPD supervises only its
own responses, making such gains especially difficult to transfer. If the
student solves the problem only sometimes, it faces a \textbf{precision gap}:
it needs targeted correction of specific bad choices, whereas indiscriminate
distillation may transfer only part of the gain or suppress correct behavior
already acquired.

Conversely, when the GRPO teacher does not outperform the student, it has
little improvement to transfer. When their performance is comparable or both
fail, the teacher offers no better direction; when the student is stronger,
matching the teacher may be harmful. Standard OPD nevertheless supervises
these problems uniformly. Skipping them concentrates the training budget on
problems where transfer is possible.

Existing OPD methods adapt supervision at the problem or token level
\citep{lin2026opdvr,xu2026tip,xu2026paced,zhang2026verifydistill}. Most closely
related, Sparse-to-Dense distills from a GRPO-improved teacher using a common
pipeline across problems \citep{sang2026beyond}. In contrast, we use the joint
success of the teacher and raw student to identify each problem's learning
need and match both its KL direction and token-level supervision.

We therefore propose \textbf{Transfer-Stratified On-Policy Distillation
(TS-OPD)}. It first stratifies problems by the joint pass rates of the GRPO
teacher and student, then routes knowledge and precision gaps to matched
objectives with stratum-specific token gating, skipping problems without a
teacher advantage. We evaluate TS-OPD across two teacher lineages (Qwen3-8B
before and after GRPO), three student scales (0.6B, 1.7B, and 4B), four
distillation objectives, and teacher-free direct GRPO.

Our contributions are as follows:
\begin{enumerate}
\item We characterize and decompose the transfer gap in distillation from
GRPO-improved teachers into distinct problem-level learning needs.

\item We propose \textbf{TS-OPD}, which stratifies and selects problems by observed
teacher--student success, routes distillation objectives, and gates token-level
supervision to address knowledge and precision gaps.

\item Experiments with a GRPO-improved teacher show that TS-OPD
achieves the highest macro-average correctness and smallest benchmark-averaged
transfer gap among evaluated methods at all three student scales, while
distilling only on selected problems.
\end{enumerate}

\section{Related Work}
\label{sec:related}

\paragraph{RL-improved teachers.}
DeepSeek-R1 distills an RL-trained model by fine-tuning on its responses
\citep{guo2025deepseekr1}. Sparse-to-Dense \citep{sang2026beyond}
applies GRPO \citep{shao2024deepseekmath} to a larger teacher, then a
forward-KL warmup and reverse-KL OPD uniformly across problems, and shows that
the RL-improved teacher yields a student stronger than direct GRPO; it compares
final students, not how much of the teacher's GRPO gain is inherited, or where. Direct-OPD instead transfers the
RL-induced policy shift, using the post-RL to pre-RL teacher log-ratio as a
dense reward \citep{feng2026directopd}, while
\citet{li2026rethinking} find that OPD requires compatible thinking patterns
and teacher capabilities new to the student
\citep{li2025small,zhang2025law,yue2025does}.

\paragraph{On-policy objectives.}
GKD trains the student on its own samples and admits divergences beyond forward
KL \citep{agarwal2024opd}, and MiniLLM optimizes reverse KL on-policy
\citep{gu2024minillm}. AKL, ToDi, EOPD, and TrOPD combine forward and reverse
KL through adaptive weights, per-token probability ratios, teacher entropy, or
token-level trust regions
\citep{wu2025akl,jung2025todi,jin2026entropy,xing2026trust}; none chooses the
direction from per-problem teacher and student performance
\citep{ko2024distillm,ko2025distillm2,wang2025abkd}.

\paragraph{Selective distillation.}
PACED weights each problem by $p(1-p)$ of the student pass rate $p$ and applies
a forward-then-reverse KL schedule to all problems \citep{xu2026paced}; TREK
applies forward KL on verified teacher proposals for prompts the student rarely
solves, then returns to GRPO \citep{xu2026trek,yan2025luffy}; TGOPD routes each prompt to OPD
or GRPO using verifier-scored teacher probes \citep{zhang2026verifydistill}.
Within responses, OPDVR gates the distillation reward by trajectory correctness
\citep{lin2026opdvr}, and TIP selects tokens by student entropy and
teacher--student divergence \citep{xu2026tip,wang2025beyond8020,lin2024rho1}.
Unlike these student-only or teacher-only signals, \tsopd{} uses the joint
success of the GRPO teacher and the raw student to choose each problem's
KL direction (forward for knowledge gaps, reverse for precision gaps) and adds
response-local token gating, rather than introducing problem-level selection
itself.

\section{Problem Definition and Transfer-Gap Diagnosis}
\label{sec:diagnosis}

Building on our observations on MATH, we formalize the transfer gap and identify
knowledge and precision gaps on MATH training problems. We examine these strata
alongside gain shortfalls on held-out competition benchmarks to motivate
differentiated OPD supervision.

\subsection{Evaluation Metrics and the GRPO Transfer Gap}

For each problem $x_i$, we sample $K$ responses from model $M$ under fixed
decoding settings and score them with a verifier $c_i(y)\in\{0,1\}$.
The sampled success rate is
$\widehat p_{M,i}=K^{-1}\sum_{j=1}^{K}c_i(y_{M,i,j})$.
For a test set of $N$ problems, define
\begin{equation}
\operatorname{avg@}K(M)=\frac{1}{N}\sum_{i=1}^{N}\widehat p_{M,i},
\qquad
\operatorname{pass@}K(M)=\frac{1}{N}\sum_{i=1}^{N}\mathbf 1[\widehat p_{M,i}>0].
\label{eq:metrics}
\end{equation}
\textbf{\avgk{} measures average single-response correctness}, our primary
performance metric; \textbf{\passk{} measures solution coverage within the
sampling budget} \citep{chen2021evaluating}.

Let $T_0$ and $T_1$ denote the teacher before and after GRPO, $S_0$ the raw
student, and $S_{\mathrm{OPD}}$ the student distilled from $T_1$. With test
accuracy $a(M)=100\operatorname{avg@}K(M)$, define the teacher and student gains
and the \textbf{transfer gap}:
\begin{equation}
\Delta_T=a(T_1)-a(T_0),\qquad
\Delta_S=a(S_{\mathrm{OPD}})-a(S_0),\qquad
G_{\mathrm{transfer}}=\Delta_T-\Delta_S.
\label{eq:transfer}
\end{equation}
All three quantities are measured in percentage points (pp). A positive transfer
gap means that the student's gain falls short of the teacher's GRPO gain,
not that their final accuracies differ by this amount.
Here $\Delta_S$ is the total student improvement from $S_0$, not the difference
between students distilled from $T_1$ and $T_0$. It may include knowledge already
present in $T_0$, so the gap does not isolate the loss of GRPO-acquired knowledge.
With the teacher and student initialization fixed, our goal is to reduce this
gap and outperform $S_{\mathrm{GRPO}}$, trained directly with GRPO from the same $S_0$.

\subsection{Two Learning Gaps on Training Problems}

We hypothesize that incomplete transfer partly reflects \textbf{supervision
mismatched to problem-specific learning needs}.
A stronger teacher need not outperform the student on every problem.
Where it does, the guidance needed depends on whether the student can already
generate a correct response.
Before distillation, we therefore sample
$K_{\mathrm{cal}}=8$ responses per training problem from both $T_1$ and $S_0$
and classify their joint success rates.

\textbf{A: knowledge gap}, $\widehat p_{S_0,i}=0<\widehat p_{T_1,i}$.
The student needs to discover correct choices absent from its sampled rollouts,
not merely increase the frequency of existing correct responses.
Zero observed success at a finite budget does not imply zero ability.

\textbf{C: precision gap}, $0<\widehat p_{S_0,i}<\widehat p_{T_1,i}$.
The student already produces correct responses but less reliably, calling for
targeted correction while preserving existing correct behavior.

\textbf{P: student advantage}, $\widehat p_{S_0,i}>\widehat p_{T_1,i}$;
\textbf{N: no observed gap}, equal rates, including joint failures.
Neither shows an observed teacher advantage, so neither is prioritized for distillation.
Let $\mathcal D_{\mathrm{train}}$ be the training-problem index set, with subsets
$\mathcal D_A$ and $\mathcal D_C$ for the knowledge and precision gaps.
Using $|\cdot|$ for set cardinality and $\mathbf 1[\cdot]$ for the indicator
function, define the \textbf{teacher-advantage problem fraction}:
\begin{equation}
r_{\mathrm{adv}}=\frac{|\mathcal D_A|+|\mathcal D_C|}{|\mathcal D_{\mathrm{train}}|}
=\frac{1}{|\mathcal D_{\mathrm{train}}|}\sum_{i\in\mathcal D_{\mathrm{train}}}
\mathbf 1[\widehat p_{T_1,i}>\widehat p_{S_0,i}].
\label{eq:advantage-fraction}
\end{equation}
This is the fraction of problems with an observed teacher advantage before
training, not its magnitude. All four stratum fractions use the full training
set as the denominator and sum to one.

\subsection{Training Strata and Test Gain Shortfalls}

We use Qwen3-8B before and after GRPO as the teacher and Qwen3-0.6B/1.7B as
students \citep{yang2025qwen3}. Stratification uses 7{,}496 MATH training problems
\citep{hendrycks2021math}, with eight responses per problem from each of $T_1$
and $S_0$; test diagnosis uses AIME24/25 and HMMT-Feb/Nov25, totaling 120 problems
with 16 responses per model per problem
\citep{aopsAimeArchive,hmmt2025february,hmmt2025november}.

\begin{table}[!htbp]
\centering
\small
\setlength{\tabcolsep}{5pt}
\caption{Training strata and test gain shortfalls. (a) Fractions of all 7{,}496
MATH training problems, classified using eight responses per model before
training. A/C denote knowledge/precision gaps; P/N denote student advantage/no
observed gap. (b) Test avg@16 accuracy on AIME24/25 and HMMT-Feb/Nov25 (120 problems),
with gains and transfer gaps in pp. Both students share the Qwen3-8B teacher
checkpoints; $S_{\mathrm{OPD}}$ uses standard reverse-KL OPD from $S_0$ with $T_1$.
Values are rounded independently to two decimal places.}
\label{tab:diagnosis}
\begin{tabular*}{\linewidth}{@{}l@{\extracolsep{\fill}}rrrrr@{}}
\toprule
\multicolumn{6}{@{}l@{}}{\textit{(a) Training-problem fractions (\%); teacher $T_1$, student $S_0$}} \\
Student & A & C & P & N & \textbf{A+C: $\boldsymbol{r_{\mathrm{adv}}}$} \\
0.6B & 23.24 & 53.92 & 3.55 & 19.29 & \textbf{77.16} \\
1.7B & 7.39 & 36.13 & 7.68 & 48.80 & \textbf{43.52} \\
\midrule
\multicolumn{6}{@{}l@{}}{\textit{(b) Test accuracy and gain shortfalls}; $G_{\mathrm{transfer}}=\Delta_T-\Delta_S$ (pp)} \\
\multicolumn{6}{@{}l@{}}{Shared teacher: $a(T_0)=15.21\%$, $a(T_1)=41.15\%$, $\Delta_T=25.94$ pp} \\
\multicolumn{2}{@{}l}{Student} & $a(S_0)$ (\%) & $a(S_{\mathrm{OPD}})$ (\%) & $\Delta_S$ (pp) & \textbf{$\boldsymbol{G_{\mathrm{transfer}}}$ (pp)} \\
\multicolumn{2}{@{}l}{0.6B} & 1.04 & 8.91 & 7.87 & \textbf{18.08} \\
\multicolumn{2}{@{}l}{1.7B} & 6.15 & 19.58 & 13.44 & \textbf{12.50} \\
\bottomrule
\end{tabular*}
\end{table}

Table~\ref{tab:diagnosis}(a) shows teacher-advantage fractions of 77.16\% and 43.52\%,
while panel (b) reports the corresponding test gain shortfalls of 18.08 and
12.50 pp for the 0.6B and 1.7B students, respectively.
The larger fraction accompanies the larger shortfall, but student scale and
stratum distribution both vary. The ordering also reverses across all eight
test sets, where macro-average shortfalls are 5.49 and 6.14 pp.
We therefore use the strata to describe supervision needs, not as an
established predictor of test gain shortfalls.

\subsection{Implications for OPD Supervision}

For A, correct responses may be rare in student rollouts. Because OPD queries
the teacher along student-generated prefixes \citep{agarwal2024opd,gu2024minillm},
guidance along complete correct solutions may be limited. Feedback at incorrect
prefixes remains available but may not suffice to reach a correct solution,
motivating guidance toward teacher-preferred continuations the student rarely selects.

For C, uniform supervision may miss decisive errors and spend limited budget
on P and N, leaving learnable problems undertrained. This motivates targeted
correction while preserving existing correct behavior.
These mechanism hypotheses
require fixed-budget stratification controls and token-level analysis.

We therefore prioritize A and C with supervision matched to knowledge
acquisition and precision improvement, deferring teacher matching on P and N.
Strata are assigned before training from joint teacher--student samples;
rollouts remain student-generated.

\section{Transfer-Stratified On-Policy Distillation}
\label{sec:method}

\begin{figure}[t]
\centering
\includegraphics[width=0.85\linewidth]{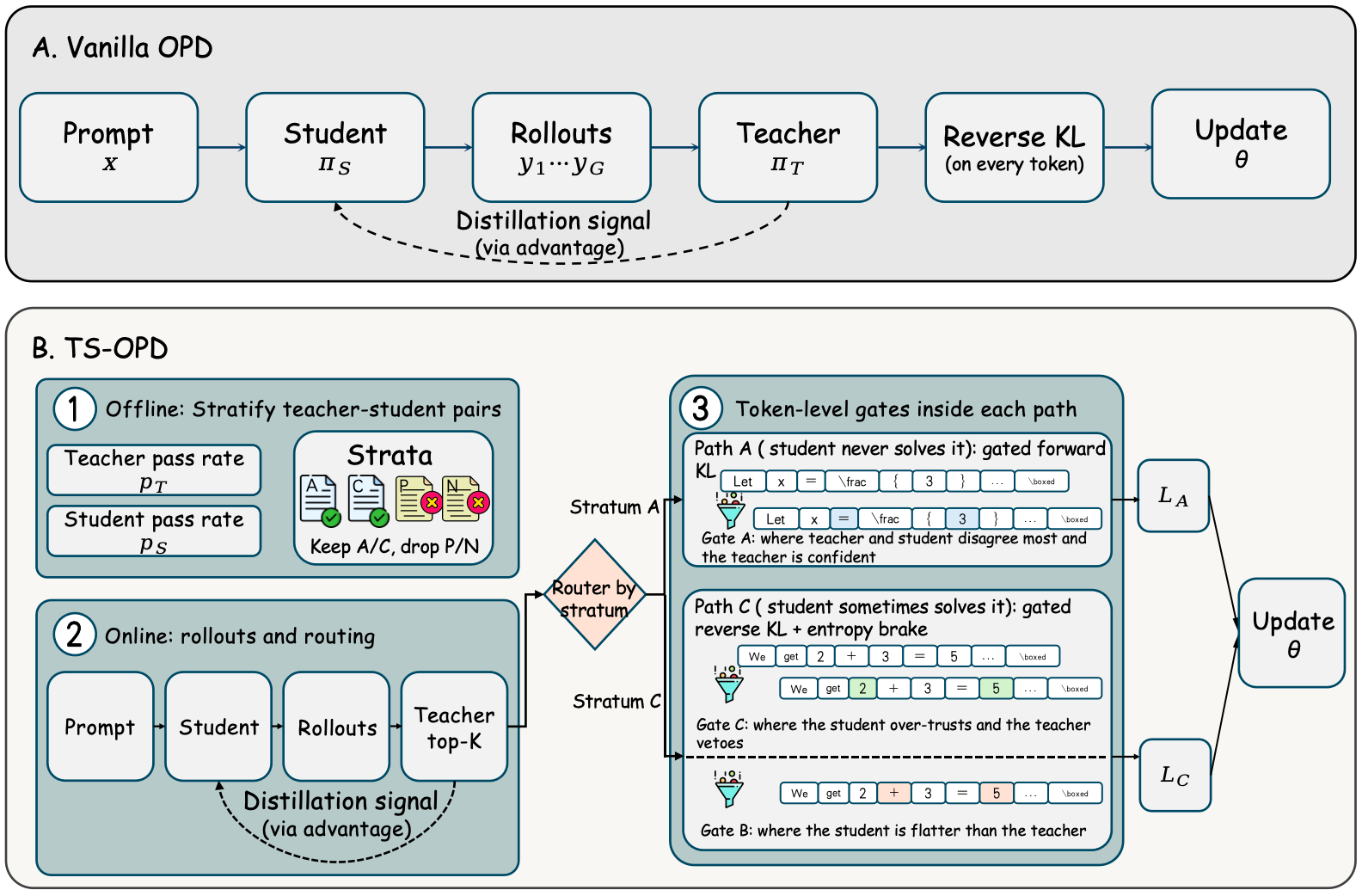}
\caption{Overview of \tsopd{}. Before training, each problem is labeled by the
sampled success of the teacher and the student. Problem level: P and N problems
are skipped, A problems use forward KL, and C problems use reverse KL. Token
level: gates select positions within a response, and C adds an entropy brake.}
\label{fig:overview}
\end{figure}

\tsopd{} labels training problems by the teacher and student success observed
before training and works at two levels (Figure~\ref{fig:overview}). At the
\textbf{problem level}, it trains only on A and C problems (\textbf{filtering})
and applies forward KL to A and reverse KL to C (\textbf{routing}). At the
\textbf{token level}, it updates only selected positions within each response
and adds an entropy brake on C (\textbf{gating}). \tsro{} keeps the problem
level and drops the token level; \tsopd{} uses both.

\subsection{Strata and KL Direction}
\label{sec:strata-kl}

\paragraph{Strata.} We use the labels $\mathcal D_A$ and $\mathcal D_C$ from
Section~3, computed once per teacher--student pair from $K_{\mathrm{cal}}=8$
samples of $T_1$ and $S_0$ and fixed during training. Filtering keeps only
$\mathcal D_+=\mathcal D_A\cup\mathcal D_C$.

\paragraph{Notation.} For a training problem $x$, the student generates a
response $y=(y_1,\ldots,y_{|y|})$ of length $|y|$. The prefix before the $t$-th
token is $s_t=(x,y_1,\ldots,y_{t-1})$. At $s_t$, the student and the teacher
($T_1$) assign next-token probabilities $\pi_S(v\mid s_t)$ and
$\pi_T(v\mid s_t)$ to each token $v$ in the vocabulary $\mathcal V$. OPD pulls
$\pi_S$ toward $\pi_T$ on these prefixes \citep{agarwal2024opd,gu2024minillm}.

\paragraph{Why forward KL for A and reverse KL for C.} At one position (we drop
$s_t$), reverse KL is $D_{\mathrm R}=\sum_v\pi_S(v)\log\frac{\pi_S(v)}{\pi_T(v)}$
and forward KL is $D_{\mathrm F}=\sum_v\pi_T(v)\log\frac{\pi_T(v)}{\pi_S(v)}$.
Let $z_v$ be the student's logit for token $v$, so $\pi_S=\mathrm{softmax}(z)$.
The gradients are
\begin{equation}
\frac{\partial D_{\mathrm R}}{\partial z_v}=\pi_S(v)\Bigl(\log\frac{\pi_S(v)}{\pi_T(v)}-D_{\mathrm R}\Bigr),
\qquad
\frac{\partial D_{\mathrm F}}{\partial z_v}=\pi_S(v)-\pi_T(v).
\label{eq:kl}
\end{equation}
Training moves $z_v$ against its gradient. For a token the teacher prefers but
the student almost never picks ($\pi_S(v)\approx0$), the reverse-KL gradient
carries the factor $\pi_S(v)$ and barely moves it, while the forward-KL
gradient is close to $-\pi_T(v)$ and raises it. A problems need behavior that
has not appeared in the student's responses, so we use forward KL. C problems
only need to drop choices the teacher rejects, so we use reverse KL
\citep{jung2025todi}. This choice per stratum is the routing. This fixed-prefix argument only motivates the loss; it
does not describe the gated update or guarantee a correct response
\citep{sheng2026ier}.

\paragraph{Token cells and position selection.} The token level (gating) works
on a compressed teacher distribution. At each position we keep the
teacher's top $K_{\mathrm{tok}}=512$ tokens plus one tail cell for the rest,
indexed by $j$ in decreasing teacher probability. $P_{t,j}$ and $Q_{t,j}$ are the teacher and student probabilities
of cell $j$ at position $t$. Both layers below select positions within a
response. Let $\qt_\tau(u)$ be the $\tau$-quantile of a score $u$ over the
positions of one response (so long responses cannot take most selections); with
$\tau=0.8$, a position passes when its score is in the top 20\% of its response.

\subsection{A Layer: Gated Forward KL}
\label{sec:layer-a}

Two kinds of positions help little on A problems. Where the student already
matches the teacher, there is nothing to learn; where the teacher spreads its
probability over many tokens, it cannot teach a definite choice. We
measure disagreement by total variation
$\mathrm{TV}_t=\frac12\sum_j|P_{t,j}-Q_{t,j}|\in[0,1]$, and teacher confidence
by $\mathrm{mass@}n(t)=\sum_{j=1}^{n}P_{t,j}$, the probability the teacher puts
on its top $n$ tokens. We keep a position only if both conditions hold, and the
A loss averages the forward KL on the cells,
$D(P_t\|Q_t)=\sum_j P_{t,j}\log(P_{t,j}/Q_{t,j})$, over the kept positions:
\begin{equation}
m^A_t=\mathbf 1\bigl[\mathrm{TV}_t\ge\qt_\tau(\mathrm{TV})\bigr]\cdot\mathbf 1\bigl[\mathrm{mass@}n(t)>0.9\bigr],
\qquad
\mathcal L^A=\frac{\sum_t m^A_t\,D(P_t\|Q_t)}{\sum_t m^A_t}.
\label{eq:gate-a}
\end{equation}
Here $\mathbf 1[\cdot]$ is 1 if the condition holds and 0 otherwise. The first
condition keeps the 20\% of positions with the largest disagreement. The second
requires the teacher to put over 90\% of its probability on a few candidates,
that is, to be confident. We do not use $n=1$ because a reasoning step often
has several equivalent forms; we set $n=\max(\lfloor\ln K_{\mathrm{tok}}\rfloor,5)=6$.

\subsection{C Layer: Gated Reverse KL and Entropy Brake}
\label{sec:layer-c}

On C problems, we only correct sampled tokens $y_t$ that the teacher rejects
and leave the rest unchanged:
\begin{equation}
r_t=\log\frac{\pi_S(y_t\mid s_t)}{\pi_T(y_t\mid s_t)},\quad
\rho_t=\max\Bigl(0,\,1-\frac{\pi_T(y_t\mid s_t)}{\pi_S(y_t\mid s_t)}\Bigr),\quad
m^C_t=\mathbf 1\bigl[\rho_t\ge\qt_\tau(\rho)\bigr]\cdot\mathbf 1[\rho_t>0].
\label{eq:gate-c}
\end{equation}
$r_t>0$ means the student is more confident in $y_t$ than the teacher;
$\rho_t\in[0,1)$ is 0 when the teacher gives $y_t$ at least the student's
probability and grows toward 1 as the teacher's falls below it.
$m^C_t$ keeps the top 20\% of positions with $\rho_t>0$; the second factor is
needed because over half of the $\rho_t$ are 0, so the quantile can be 0.

\paragraph{Lowering reverse KL with sampled tokens.} Reverse KL is the average
of $\log\frac{\pi_S(v)}{\pi_T(v)}$ over tokens drawn from the student, so $r_t$
on the sampled token $y_t$ is a one-sample estimate of it. We use $r_t$ as a
penalty: the larger $r_t$, the more we lower the probability of $y_t$. This
needs only the student's own response and no correctness reward
\citep{gu2024minillm} (derivation in Appendix~\ref{app:rkl}). With
$\omega_t=-m^C_t r_t$ held constant, the reverse-KL loss is
\begin{equation}
\mathcal L^{\mathrm{RKL}}=-\frac{1}{|y|}\textstyle\sum_{t}\omega_t\,\log\pi_S(y_t\mid s_t).
\label{eq:loss-c}
\end{equation}
Unselected positions have $\omega_t=0$ and are not updated.

\paragraph{Entropy brake.} Reverse KL concentrates probability on a few tokens,
which can remove other correct solutions and lower \passk{}. Let $H_S(t)=-\sum_jQ_{t,j}\log Q_{t,j}$ and
$H_T(t)=-\sum_jP_{t,j}\log P_{t,j}$ be the student and teacher entropies. Where
$H_S-H_T$ is largest, reverse KL sharpens the student the most, so we raise the
student entropy there, at positions picked by $m^B_t$:
\begin{equation}
m^B_t=\mathbf 1\bigl[H_S(t)-H_T(t)\ge\qt_\tau(H_S-H_T)\bigr],\qquad
\mathcal L^B=-\beta\,\frac{\sum_t m^B_t H_S(t)}{\sum_t m^B_t}.
\label{eq:brake}
\end{equation}
We set $\beta$ once per pair before training so that the brake is about a
tenth of the penalty in magnitude, a heuristic scale match rather than a
gradient ratio, with 0.1 as a design choice (Appendix~\ref{app:beta}).

\paragraph{Full objective and training loop.}\label{sec:objective} The C loss is
$\mathcal L^C=\mathcal L^{\mathrm{RKL}}+\mathcal L^B$, and a response to problem $x$
uses the loss of its stratum:
\begin{equation}
\mathcal L=\mathcal L^A\,\mathbf 1[x\in\mathcal D_A]+\mathcal L^C\,\mathbf 1[x\in\mathcal D_C].
\label{eq:full}
\end{equation}
Problems outside $\mathcal D_+$ give no loss. Each step, the student answers a batch of problems from
$\mathcal D_+$ once, the teacher scores the answers, and we make one update;
gates get no gradient (Algorithm~\ref{alg:tsopd}). \tsro{} removes gating by
setting $m^A_t\equiv m^C_t\equiv1$ and $\beta=0$: $\mathcal L^A$ becomes forward
KL averaged over all positions, and $\mathcal L^C$ becomes the sampled reverse KL
on all positions.

\section{Experiments}
\label{sec:experiments}

\paragraph{Models and arms.} The teachers are $T_0$ (base-8B) and $T_1$
(GRPO-8B); the students are Qwen3-0.6B, 1.7B, and 4B \citep{yang2025qwen3}.
Every run trains for one epoch on the 7{,}496 MATH training problems
\citep{hendrycks2021math}. Direct GRPO trains the same student with GRPO and no
teacher; it is the $S_{\mathrm{GRPO}}$ that Section~3 sets out to beat. We
compare it with OPD-r and OPD-f (uniform reverse and forward KL on all
problems), and \tsro{} and \tsopd{}, which train on \AC{} only and so take about
91, 51, and 27 of the 118 OPD steps. Three controls, Direct GRPO (\AC), OPD-r
(\AC), and OPD-f (\AC), use the same problems and steps as \tsopd{}; only the
loss differs. Details are in Appendix~\ref{app:setup}.

\paragraph{Benchmarks and metrics.} We sample $K$ responses per problem on eight
benchmarks: MATH-test (5{,}000 problems, $K=1$) and MATH500 ($K=8$)
\citep{hendrycks2021math,lightman2023verify}, Minerva
\citep{lewkowycz2022minerva} and OlympiadBench \citep{he2024olympiadbench}
($K=8$), and AIME24/25 \citep{aopsAimeArchive} and HMMT-Feb/Nov25
\citep{hmmt2025february,hmmt2025november} ($K=16$). We report macro \avgk{} and
\passk{} over all eight. With $K=1$, a MATH-test problem is only right or wrong
and cannot show a change in reliability, so $\Gt$ (Section~3) and the
per-problem analysis use the other seven benchmarks (1{,}566 problems).

\paragraph{Outcome categories.} To trace gains per test problem, we compare a
model's sampled accuracy $p_0$ before and $p_1$ after training: $\tA$ if
$p_0=0<p_1$ (newly solved), $\tC$ if $0<p_0<p_1$ (more reliable), $\tP$ if
$p_0>p_1$ (worse), and $\tN$ otherwise (unchanged); ``before'' is $S_0$ for a
student and $T_0$ for the teacher. Test problems are also split into strata by
$S_0$ and $T_1$ with the rule of Section~3.

\subsection{Teacher GRPO Is Necessary but Not Sufficient}
\label{sec:exp-main}

\begin{table}[htbp]
\centering
\small
\setlength{\tabcolsep}{3.4pt}
\caption{Main results. \avgk{} and \passk{} are macro averages over the eight
benchmarks (\%). $\Gt$ is the transfer gap in percentage points, computed on
seven benchmarks. $\uparrow$/$\downarrow$: higher/lower is better. Direct GRPO
uses no teacher. In each row, bold marks the best value and underline the second
best.}
\label{tab:main}
\setlength{\aboverulesep}{0pt}\setlength{\belowrulesep}{0pt}\renewcommand{\arraystretch}{1.12}
\begin{tabular}{llrrrrrrrr|r}
\toprule
 & & \multicolumn{4}{c}{base-8B teacher} & \multicolumn{4}{c|}{GRPO-8B teacher} & \\
\cmidrule(lr){3-6}\cmidrule(lr){7-10}
 & & & & \multicolumn{2}{c}{TS-OPD} & & & \multicolumn{2}{c|}{TS-OPD} & Direct \\
\cmidrule(lr){5-6}\cmidrule(lr){9-10}
Student & Metric & OPD-r & OPD-f & routing only & full & OPD-r & OPD-f & routing only & full & GRPO \\
\midrule
0.6B & avg@$K$ $\uparrow$ & 19.61 & 19.15 & 20.60 & 21.27 & 27.00 & 26.57 & \underline{27.62} & \textbf{28.88} & 26.04 \\
 & pass@$K$ $\uparrow$ & 32.80 & 32.91 & 34.14 & 36.22 & 44.11 & 41.96 & \textbf{44.83} & \underline{44.57} & 40.46 \\
 & $G_{\mathrm{transfer}}$ $\downarrow$ & 15.10 & 15.51 & 13.84 & 13.06 & 7.39 & 8.22 & \underline{6.93} & \textbf{5.68} & 8.71 \\
\midrule
1.7B & avg@$K$ $\uparrow$ & 29.22 & 27.57 & 29.29 & 29.71 & 39.04 & 39.25 & \underline{39.84} & \textbf{40.50} & 39.53 \\
 & pass@$K$ $\uparrow$ & 44.21 & 44.90 & 48.49 & 45.76 & 56.83 & 57.57 & \underline{59.34} & \textbf{59.78} & 57.33 \\
 & $G_{\mathrm{transfer}}$ $\downarrow$ & 17.40 & 19.25 & 17.27 & 16.71 & 7.28 & 7.15 & \underline{6.43} & \textbf{5.77} & 6.94 \\
\midrule
4B & avg@$K$ $\uparrow$ & 36.07 & 36.12 & 36.53 & 36.79 & 52.27 & \underline{52.83} & 52.76 & \textbf{54.28} & 51.57 \\
 & pass@$K$ $\uparrow$ & 53.93 & 54.06 & 52.81 & 53.31 & 70.71 & \underline{72.43} & \textbf{73.68} & 71.04 & 68.44 \\
 & $G_{\mathrm{transfer}}$ $\downarrow$ & 20.06 & 19.99 & 19.53 & 19.21 & 2.65 & \underline{1.98} & 2.09 & \textbf{0.46} & 3.45 \\
\bottomrule
\end{tabular}

\end{table}

With the base-8B teacher, every OPD objective trails Direct GRPO on both metrics
at every size, and the transfer gap is 13 to 20 points (Table~\ref{tab:main}).
The teacher bounds what OPD can reach: Direct GRPO at 1.7B and 4B (39.53 and
51.57) already beats the base-8B teacher itself (38.03). With the GRPO-8B
teacher, \tsopd{} has the best \avgk{} and the smallest gap at every size. It is
2.84, 0.97, and 2.71 points above Direct GRPO, and its gap is 5.68, 5.77, and
0.46 points. \passk{} is also highest for \tsopd{} or \tsro{} at every size.
Uniform OPD does not reliably help: at 1.7B, both OPD-r and OPD-f fall below
Direct GRPO and have a larger gap. The gap nearly closes at 4B because the raw
4B student is almost as strong as $T_0$ (30.73 vs.\ 31.54).

\subsection{Only Part of the Teacher's GRPO Gain Reaches the Student}
\label{sec:exp-where}

\begin{figure}[t]
\centering
\includegraphics[width=\linewidth]{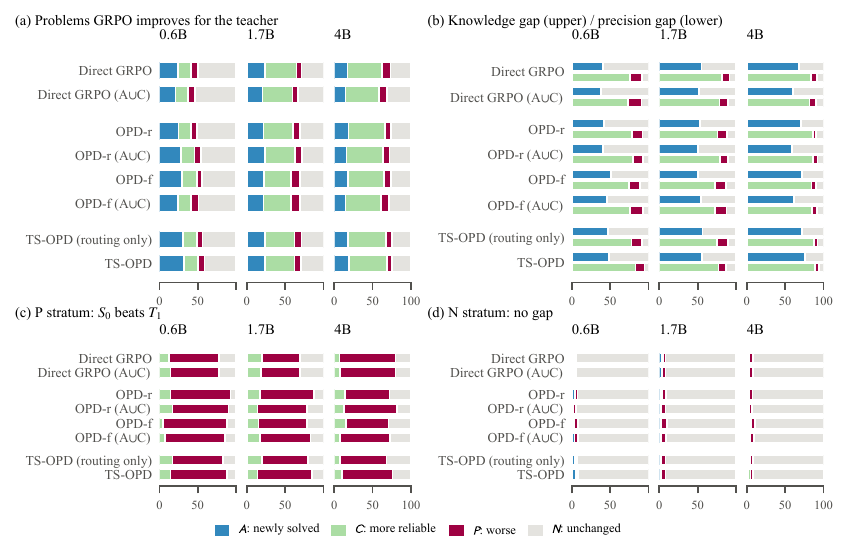}
\caption{Outcome categories of test problems in each student. (a) The 567
problems GRPO improves for the teacher. (b) Knowledge-gap (upper bar) and
precision-gap (lower bar) problems. (c, d) Problems where $T_1$ is not better
than $S_0$. Numbers: Tables~\ref{tab:app-rq3}, \ref{tab:rq4}, and~\ref{tab:rq2}.}
\label{fig:where}
\end{figure}

$\Gt$ cannot tell whether what GRPO taught the teacher reaches the student
(Section~3), so Figure~\ref{fig:where}(a) follows the 567 test problems
that GRPO improves for the teacher. \tsopd{} improves 50.1\%, 62.1\%, and 68.4\%
of them. Even at 4B, where the gap is almost zero, a third do not improve.
The teacher is what makes the difference on these problems: Direct GRPO (\AC), trained
on the same problems without a teacher, improves only 37.1\%, 59.5\%, and
58.1\%, and \tsopd{} newly solves the largest share ($\tA$) at every size.

Figure~\ref{fig:where}(b) shows where the teacher helps. Compared with Direct
GRPO, \tsopd{} newly solves 7.0, 0.4, and 7.9 points more knowledge-gap
problems. On precision-gap problems the picture is mixed: \tsopd{} is ahead at
0.6B and 4B, but at 1.7B Direct GRPO makes the most problems reliable (81.6\%)
and the fewest worse (9.7\%). This fits Section~3: GRPO can only sharpen what the
student already samples, which is what a precision gap needs, and it explains
why \tsopd{} leads Direct GRPO by only 0.97 points at 1.7B.

The loss comparisons on the same \AC{} problems support the choice of KL
direction in Section~\ref{sec:strata-kl}. Forward KL newly solves more
knowledge-gap problems than reverse KL (OPD-f (\AC) 46.2\%, 54.5\%, 61.8\% vs.\
OPD-r (\AC) 40.2\%, 50.4\%, 59.2\%), because it can raise tokens the student
rarely picks. Reverse KL makes fewer precision-gap problems worse (13.0\%,
11.0\%, 5.7\% vs.\ 16.9\%, 14.8\%, 6.5\%), because it only lowers tokens the
teacher rejects and leaves the rest of the student's solution alone. Routing
uses each loss where it helps and newly solves more knowledge-gap problems than
either loss alone (47.0\%, 56.2\%, 71.7\%). Gates then mainly help precision-gap
problems: \tsopd{} makes 5.4, 2.0, and 1.0 points more of them reliable than
\tsro{}, which matches their design of correcting only a few rejected tokens.

\subsection{Filtering Helps Only with Routing and Gating}
\label{sec:exp-components}

\begin{table}[t]
\centering
\small
\caption{Adding one component at a time: macro \avgk{} (\%) / $\Gt$ (points),
GRPO-8B teacher. All \AC{} rows use the same problems and steps.}
\label{tab:rq5}
\begin{tabular}{llccc}
\toprule
Step & Method & 0.6B & 1.7B & 4B \\
\midrule
No teacher & Direct GRPO & 26.04 / 8.71 & 39.53 / 6.94 & 51.57 / 3.45 \\
\quad + \AC{} only & Direct GRPO (\AC) & 24.91 / 9.34 & 37.82 / 8.16 & 47.48 / 4.21 \\
\midrule
Start & OPD-r & 27.00 / 7.39 & 39.04 / 7.28 & 52.27 / 2.65 \\
\quad + filtering & OPD-r (\AC) & 25.66 / 9.02 & 37.51 / 9.12 & 47.70 / 7.87 \\
Start & OPD-f & 26.57 / 8.22 & 39.25 / 7.15 & 52.83 / 1.98 \\
\quad + filtering & OPD-f (\AC) & 24.87 / 10.17 & 36.52 / 10.34 & 47.19 / 8.45 \\
\quad + routing & \tsro{} & 27.62 / 6.93 & 39.84 / 6.43 & 52.76 / 2.09 \\
\quad + gating & \tsopd{} & 28.88 / 5.68 & 40.50 / 5.77 & 54.28 / 0.46 \\
\bottomrule
\end{tabular}

\end{table}

Figure~\ref{fig:where}(c, d) checks whether the skipped P and N problems matter. On
the P stratum,
where $S_0$ beats $T_1$, every method makes about half or more of the problems
worse, including Direct GRPO without a teacher (49.2\% to 72.8\%). These
problems were picked by samples on which $S_0$ did well, so much of this drop is
regression to the mean; training on \AC{} only does not change it consistently. Gates lose more of them than \tsro{} (73.2\% vs.\ 65.9\% at
0.6B), likely because they pull the student harder toward the teacher. On the larger N stratum, more than 89\% stay unchanged for every
method, and skipping changes the share that gets worse by at most 1.3 points.

Skipping, however, is not free. Table~\ref{tab:rq5} shows that filtering alone
hurts: with the same problems and steps as \tsopd{}, OPD-r (\AC) and OPD-f
(\AC) lose 1.3 to 5.6 points of \avgk{}, and their gap grows at every size.
Figure~\ref{fig:where}(b) locates the loss. On knowledge-gap problems, filtering
lowers the newly solved share for all three losses at 0.6B and 4B, by 7.9 to
11.9 points at 4B, where the \AC{} runs take only 27 of 118 steps. On
precision-gap problems, it does not hurt OPD: the share made reliable stays or
rises (by 0.3 to 2.9 points), since the remaining updates fall on problems the
student already partly solves. Direct GRPO (\AC) drops on both, because the student
never succeeds on A problems and GRPO gets no reward there. Acquiring a new
solution thus needs more supervision than sharpening an existing one.

Routing makes up for this. Forward KL on A problems raises the newly solved share
of knowledge-gap problems by 6.8, 5.8, and 12.5 points over OPD-r (\AC), and
reaches the level of OPD on all problems at 1.7B and 4B with fewer steps. As a
result, routing lowers the gap by 2.09, 2.69, and 5.78 points, and \tsro{} beats
OPD-r on all problems in both \avgk{} and $\Gt$. Gating lowers the gap by another
1.25, 0.66, and 1.63 points, mainly through precision-gap problems; the 0.6B
ablations in Table~\ref{tab:ablation} show that each gate setting and the brake
matter.

\section{Conclusion}

Teacher improvement and student transfer are different objects. In mathematical
reasoning, a GRPO-improved teacher can provide useful dense supervision, but
only when OPD distinguishes acquisition gaps from consolidation gaps and avoids
spending budget on no-gap or student-advantage problems. \tsopd{} implements
this transfer-aware view with problem-level routing, response-local token
gates, and an entropy brake. Across two teacher lineages and three student
sizes, the results show that the routed objective is most useful with a
GRPO-improved teacher and that its gains depend on the full design. The broader
lesson is that stronger teachers should be evaluated not only by endpoint
accuracy, but by how their improvements decompose into teachable strata for the
student.

\section*{Limitations}

\textsf{\textbf{Scope.}} All experiments use mathematical reasoning with verifiable
answers, one model family (Qwen3), one teacher size (8B), and one RL algorithm
(GRPO) for improving the teacher. The students go up to 4B. We have not tested
whether the same strata and routing carry over to other domains, to students
close to the teacher in size, or to teachers improved by other RL methods.

\textsf{\textbf{Statistical strength.}} Most runs use a single training seed, so small
differences may not be
robust. We therefore read the per-problem outcome categories as
trends rather than exact rates.

\textsf{\textbf{Strata and budget.}} The strata come from $K_{\mathrm{cal}}=8$
samples per training problem, taken once before training. With this budget, a
problem with no observed student success may still be solvable, and a problem
can move between strata as the student improves, but \tsopd{} does not relabel
it.

\textsf{\textbf{Design choices and cost.}} The hyperparameter ablations use only the
0.6B student with the GRPO-8B teacher.
Screening needs $2K_{\mathrm{cal}}$ sampled responses per training problem
before training, and every update needs the teacher's top $K_{\mathrm{tok}}$
probabilities at each position. Finally, $\Gt$ measures a gain shortfall; it
does not isolate how much GRPO-acquired knowledge is lost (Section~3).

\bibliography{paper_refs}

\begin{thebibliography}{35}
\providecommand{\natexlab}[1]{#1}
\providecommand{\url}[1]{\texttt{#1}}
\expandafter\ifx\csname urlstyle\endcsname\relax
  \providecommand{\doi}[1]{doi: #1}\else
  \providecommand{\doi}{doi: \begingroup \urlstyle{rm}\Url}\fi

\bibitem[Agarwal et~al.(2024)Agarwal, Vieillard, Zhou, Stanczyk, Ramos, Geist,
  and Bachem]{agarwal2024opd}
Rishabh Agarwal, Nino Vieillard, Yongchao Zhou, Piotr Stanczyk, Sabela Ramos,
  Matthieu Geist, and Olivier Bachem.
\newblock On-policy distillation of language models: Learning from
  self-generated mistakes, 2024.
\newblock URL \url{https://arxiv.org/abs/2306.13649}.

\bibitem[{Art of Problem Solving}(n.d.)]{aopsAimeArchive}
{Art of Problem Solving}.
\newblock {AIME} problems and solutions.
\newblock AoPS Wiki, n.d.
\newblock URL
  \url{https://artofproblemsolving.com/wiki/index.php/AIME_Problems_and_Solutions}.
\newblock Community-maintained problem archive. Accessed September 23, 2026.

\bibitem[Chen et~al.(2021)Chen, Tworek, Jun, Yuan, de~Oliveira~Pinto, Kaplan,
  Edwards, Burda, Joseph, Brockman, Ray, Puri, Krueger, Petrov, Khlaaf, Sastry,
  Mishkin, Chan, Gray, Ryder, Pavlov, Power, Kaiser, Bavarian, Winter, Tillet,
  Such, Cummings, Plappert, Chantzis, Barnes, Herbert-Voss, Guss, Nichol,
  Paino, Tezak, Tang, Babuschkin, Balaji, Jain, Saunders, Hesse, Carr, Leike,
  Achiam, Misra, Morikawa, Radford, Knight, Brundage, Murati, Mayer, Welinder,
  McGrew, Amodei, McCandlish, Sutskever, and Zaremba]{chen2021evaluating}
Mark Chen, Jerry Tworek, Heewoo Jun, Qiming Yuan, Henrique~Ponde
  de~Oliveira~Pinto, Jared Kaplan, Harri Edwards, Yuri Burda, Nicholas Joseph,
  Greg Brockman, Alex Ray, Raul Puri, Gretchen Krueger, Michael Petrov, Heidy
  Khlaaf, Girish Sastry, Pamela Mishkin, Brooke Chan, Scott Gray, Nick Ryder,
  Mikhail Pavlov, Alethea Power, Lukasz Kaiser, Mohammad Bavarian, Clemens
  Winter, Philippe Tillet, Felipe~Petroski Such, Dave Cummings, Matthias
  Plappert, Fotios Chantzis, Elizabeth Barnes, Ariel Herbert-Voss,
  William~Hebgen Guss, Alex Nichol, Alex Paino, Nikolas Tezak, Jie Tang, Igor
  Babuschkin, Suchir Balaji, Shantanu Jain, William Saunders, Christopher
  Hesse, Andrew~N. Carr, Jan Leike, Josh Achiam, Vedant Misra, Evan Morikawa,
  Alec Radford, Matthew Knight, Miles Brundage, Mira Murati, Katie Mayer, Peter
  Welinder, Bob McGrew, Dario Amodei, Sam McCandlish, Ilya Sutskever, and
  Wojciech Zaremba.
\newblock Evaluating large language models trained on code, 2021.
\newblock URL \url{https://arxiv.org/abs/2107.03374}.

\bibitem[Feng et~al.(2026)Feng, ang Gao, Chi, Wu, Zhang, Jiang, He, Ma, Zhang,
  and Zhou]{feng2026directopd}
Shiyuan Feng, Huan ang Gao, Haohan Chi, Hanlin Wu, Zhilong Zhang, Zheng Jiang,
  Bingxiang He, Wei-Ying Ma, Ya-Qin Zhang, and Hao Zhou.
\newblock Weak-to-strong generalization via direct on-policy distillation,
  2026.
\newblock URL \url{https://arxiv.org/abs/2607.05394}.

\bibitem[Gu et~al.(2024)Gu, Dong, Wei, and Huang]{gu2024minillm}
Yuxian Gu, Li~Dong, Furu Wei, and Minlie Huang.
\newblock Mini{LLM}: Knowledge distillation of large language models.
\newblock In \emph{The Twelfth International Conference on Learning
  Representations}, 2024.
\newblock URL \url{https://openreview.net/forum?id=5h0qf7IBZZ}.

\bibitem[Guo et~al.(2025)Guo, Yang, Zhang, Song, Wang, Zhu, Xu, Zhang, Ma, Bi,
  Zhang, Yu, Wu, Wu, Gou, Shao, Li, Gao, Liu, Xue, Wang, Wu, Feng, Lu, Zhao,
  Deng, Ruan, Dai, Chen, Ji, Li, Lin, Dai, Luo, Hao, Chen, Li, Zhang, Xu, Ding,
  Gao, Qu, Li, Guo, Li, Chen, Yuan, Tu, Qiu, Li, Cai, Ni, Liang, Chen, Dong,
  Hu, You, Gao, Guan, Huang, Yu, Wang, Zhang, Zhao, Wang, Zhang, Xu, Xia,
  Zhang, Zhang, Tang, Zhou, Li, Wang, Li, Tian, Huang, Zhang, Wang, Chen, Du,
  Ge, Zhang, Pan, Wang, Chen, Jin, Chen, Lu, Zhou, Chen, Ye, Wang, Yu, Zhou,
  Pan, Li, Zhou, Wu, Yun, Pei, Sun, Wang, Zeng, Liu, Liang, Gao, Yu, Zhang,
  Xiao, An, Liu, Wang, Chen, Nie, Cheng, Liu, Xie, Liu, Yang, Li, Su, Lin, Li,
  Jin, Shen, Chen, Sun, Wang, Song, Zhou, Wang, Shan, Li, Wang, Wei, Zhang, Xu,
  Li, Zhao, Sun, Wang, Yu, Zhang, Shi, Xiong, He, Piao, Wang, Tan, Ma, Liu,
  Guo, Ou, Wang, Gong, Zou, He, Xiong, Luo, You, Liu, Zhou, Zhu, Huang, Li,
  Zheng, Zhu, Ma, Tang, Zha, Yan, Ren, Ren, Sha, Fu, Xu, Xie, Zhang, Hao, Ma,
  Yan, Wu, Gu, Zhu, Liu, Li, Xie, Song, Pan, Huang, Xu, Zhang, and
  Zhang]{guo2025deepseekr1}
Daya Guo, Dejian Yang, Haowei Zhang, Junxiao Song, Peiyi Wang, Qihao Zhu,
  Runxin Xu, Ruoyu Zhang, Shirong Ma, Xiao Bi, Xiaokang Zhang, Xingkai Yu,
  Yu~Wu, Z.~F. Wu, Zhibin Gou, Zhihong Shao, Zhuoshu Li, Ziyi Gao, Aixin Liu,
  Bing Xue, Bingxuan Wang, Bochao Wu, Bei Feng, Chengda Lu, Chenggang Zhao,
  Chengqi Deng, Chong Ruan, Damai Dai, Deli Chen, Dongjie Ji, Erhang Li,
  Fangyun Lin, Fucong Dai, Fuli Luo, Guangbo Hao, Guanting Chen, Guowei Li,
  H.~Zhang, Hanwei Xu, Honghui Ding, Huazuo Gao, Hui Qu, Hui Li, Jianzhong Guo,
  Jiashi Li, Jingchang Chen, Jingyang Yuan, Jinhao Tu, Junjie Qiu, Junlong Li,
  J.~L. Cai, Jiaqi Ni, Jian Liang, Jin Chen, Kai Dong, Kai Hu, Kaichao You,
  Kaige Gao, Kang Guan, Kexin Huang, Kuai Yu, Lean Wang, Lecong Zhang, Liang
  Zhao, Litong Wang, Liyue Zhang, Lei Xu, Leyi Xia, Mingchuan Zhang, Minghua
  Zhang, Minghui Tang, Mingxu Zhou, Meng Li, Miaojun Wang, Mingming Li, Ning
  Tian, Panpan Huang, Peng Zhang, Qiancheng Wang, Qinyu Chen, Qiushi Du, Ruiqi
  Ge, Ruisong Zhang, Ruizhe Pan, Runji Wang, R.~J. Chen, R.~L. Jin, Ruyi Chen,
  Shanghao Lu, Shangyan Zhou, Shanhuang Chen, Shengfeng Ye, Shiyu Wang,
  Shuiping Yu, Shunfeng Zhou, Shuting Pan, S.~S. Li, Shuang Zhou, Shaoqing Wu,
  Tao Yun, Tian Pei, Tianyu Sun, T.~Wang, Wangding Zeng, Wen Liu, Wenfeng
  Liang, Wenjun Gao, Wenqin Yu, Wentao Zhang, W.~L. Xiao, Wei An, Xiaodong Liu,
  Xiaohan Wang, Xiaokang Chen, Xiaotao Nie, Xin Cheng, Xin Liu, Xin Xie,
  Xingchao Liu, Xinyu Yang, Xinyuan Li, Xuecheng Su, Xuheng Lin, X.~Q. Li,
  Xiangyue Jin, Xiaojin Shen, Xiaosha Chen, Xiaowen Sun, Xiaoxiang Wang, Xinnan
  Song, Xinyi Zhou, Xianzu Wang, Xinxia Shan, Y.~K. Li, Y.~Q. Wang, Y.~X. Wei,
  Yang Zhang, Yanhong Xu, Yao Li, Yao Zhao, Yaofeng Sun, Yaohui Wang, Yi~Yu,
  Yichao Zhang, Yifan Shi, Yiliang Xiong, Ying He, Yishi Piao, Yisong Wang,
  Yixuan Tan, Yiyang Ma, Yiyuan Liu, Yongqiang Guo, Yuan Ou, Yuduan Wang, Yue
  Gong, Yuheng Zou, Yujia He, Yunfan Xiong, Yuxiang Luo, Yuxiang You, Yuxuan
  Liu, Yuyang Zhou, Y.~X. Zhu, Yanping Huang, Yaohui Li, Yi~Zheng, Yuchen Zhu,
  Yunxian Ma, Ying Tang, Yukun Zha, Yuting Yan, Z.~Z. Ren, Zehui Ren, Zhangli
  Sha, Zhe Fu, Zhean Xu, Zhenda Xie, Zhengyan Zhang, Zhewen Hao, Zhicheng Ma,
  Zhigang Yan, Zhiyu Wu, Zihui Gu, Zijia Zhu, Zijun Liu, Zilin Li, Ziwei Xie,
  Ziyang Song, Zizheng Pan, Zhen Huang, Zhipeng Xu, Zhongyu Zhang, and Zhen
  Zhang.
\newblock Deepseek-r1 incentivizes reasoning in llms through reinforcement
  learning.
\newblock \emph{Nature}, 645\penalty0 (8081):\penalty0 633--638, September
  2025.
\newblock ISSN 1476-4687.
\newblock \doi{10.1038/s41586-025-09422-z}.
\newblock URL \url{http://dx.doi.org/10.1038/s41586-025-09422-z}.

\bibitem[He et~al.(2024)He, Luo, Bai, Hu, Thai, Shen, Hu, Han, Huang, Zhang,
  Liu, Qi, Liu, and Sun]{he2024olympiadbench}
Chaoqun He, Renjie Luo, Yuzhuo Bai, Shengding Hu, Zhen~Leng Thai, Junhao Shen,
  Jinyi Hu, Xu~Han, Yujie Huang, Yuxiang Zhang, Jie Liu, Lei Qi, Zhiyuan Liu,
  and Maosong Sun.
\newblock Olympiadbench: A challenging benchmark for promoting agi with
  olympiad-level bilingual multimodal scientific problems, 2024.
\newblock URL \url{https://arxiv.org/abs/2402.14008}.

\bibitem[Hendrycks et~al.(2021)Hendrycks, Burns, Kadavath, Arora, Basart, Tang,
  Song, and Steinhardt]{hendrycks2021math}
Dan Hendrycks, Collin Burns, Saurav Kadavath, Akul Arora, Steven Basart, Eric
  Tang, Dawn Song, and Jacob Steinhardt.
\newblock Measuring mathematical problem solving with the math dataset, 2021.
\newblock URL \url{https://arxiv.org/abs/2103.03874}.

\bibitem[{HMMT}(2025{\natexlab{a}})]{hmmt2025february}
{HMMT}.
\newblock Archive of february 2025, 2025{\natexlab{a}}.
\newblock URL \url{https://www.hmmt.org/www/archive/282}.
\newblock Official problem and solution archive. Accessed September 23, 2026.

\bibitem[{HMMT}(2025{\natexlab{b}})]{hmmt2025november}
{HMMT}.
\newblock Archive of november 2025, 2025{\natexlab{b}}.
\newblock URL \url{https://www.hmmt.org/www/archive/291}.
\newblock Official problem and solution archive. Accessed September 23, 2026.

\bibitem[Jin et~al.(2026)Jin, Min, Yang, Wei, Zhou, Kadhe, Baracaldo, and
  Lee]{jin2026entropy}
Woogyeol Jin, Taywon Min, Yongjin Yang, Dennis Wei, Yi~Zhou, Swanand~Ravindra
  Kadhe, Nathalie Baracaldo, and Kimin Lee.
\newblock Entropy-aware on-policy distillation of language models, 2026.
\newblock URL \url{https://arxiv.org/abs/2603.07079}.

\bibitem[Jung et~al.(2025)Jung, Yoon, Kim, and Lee]{jung2025todi}
Seongryong Jung, Suwan Yoon, DongGeon Kim, and Hwanhee Lee.
\newblock Todi: Token-wise distillation via fine-grained divergence control,
  2025.
\newblock URL \url{https://arxiv.org/abs/2505.16297}.

\bibitem[Ko et~al.(2024)Ko, Kim, Chen, and Yun]{ko2024distillm}
Jongwoo Ko, Sungnyun Kim, Tianyi Chen, and Se-Young Yun.
\newblock Distillm: Towards streamlined distillation for large language models,
  2024.
\newblock URL \url{https://arxiv.org/abs/2402.03898}.

\bibitem[Ko et~al.(2025)Ko, Chen, Kim, Ding, Liang, Zharkov, and
  Yun]{ko2025distillm2}
Jongwoo Ko, Tianyi Chen, Sungnyun Kim, Tianyu Ding, Luming Liang, Ilya Zharkov,
  and Se-Young Yun.
\newblock Distillm-2: A contrastive approach boosts the distillation of llms,
  2025.
\newblock URL \url{https://arxiv.org/abs/2503.07067}.

\bibitem[Lewkowycz et~al.(2022)Lewkowycz, Andreassen, Dohan, Dyer, Michalewski,
  Ramasesh, Slone, Anil, Schlag, Gutman-Solo, Wu, Neyshabur, Gur-Ari, and
  Misra]{lewkowycz2022minerva}
Aitor Lewkowycz, Anders Andreassen, David Dohan, Ethan Dyer, Henryk
  Michalewski, Vinay Ramasesh, Ambrose Slone, Cem Anil, Imanol Schlag, Theo
  Gutman-Solo, Yuhuai Wu, Behnam Neyshabur, Guy Gur-Ari, and Vedant Misra.
\newblock Solving quantitative reasoning problems with language models, 2022.
\newblock URL \url{https://arxiv.org/abs/2206.14858}.

\bibitem[Li et~al.(2026)Li, Zuo, He, Zhang, Xiao, Qian, Yu, ang Gao, Yang, Liu,
  and Ding]{li2026rethinking}
Yaxuan Li, Yuxin Zuo, Bingxiang He, Jinqian Zhang, Chaojun Xiao, Cheng Qian,
  Tianyu Yu, Huan ang Gao, Wenkai Yang, Zhiyuan Liu, and Ning Ding.
\newblock Rethinking on-policy distillation of large language models:
  Phenomenology, mechanism, and recipe, 2026.
\newblock URL \url{https://arxiv.org/abs/2604.13016}.

\bibitem[Li et~al.(2025)Li, Yue, Xu, Jiang, Niu, Lin, Ramasubramanian, and
  Poovendran]{li2025small}
Yuetai Li, Xiang Yue, Zhangchen Xu, Fengqing Jiang, Luyao Niu, Bill~Yuchen Lin,
  Bhaskar Ramasubramanian, and Radha Poovendran.
\newblock Small models struggle to learn from strong reasoners, 2025.
\newblock URL \url{https://arxiv.org/abs/2502.12143}.

\bibitem[Lightman et~al.(2023)Lightman, Kosaraju, Burda, Edwards, Baker, Lee,
  Leike, Schulman, Sutskever, and Cobbe]{lightman2023verify}
Hunter Lightman, Vineet Kosaraju, Yura Burda, Harri Edwards, Bowen Baker, Teddy
  Lee, Jan Leike, John Schulman, Ilya Sutskever, and Karl Cobbe.
\newblock Let's verify step by step, 2023.
\newblock URL \url{https://arxiv.org/abs/2305.20050}.

\bibitem[Lin et~al.(2026)Lin, Zhao, Jiang, Rao, Li, Wang, He, and
  Huang]{lin2026opdvr}
Wenze Lin, Jiale Zhao, Xitai Jiang, Songde Rao, Yining Li, Shenzhi Wang,
  Bingxiang He, and Gao Huang.
\newblock On-policy distillation with verifiable reward, 2026.
\newblock URL \url{https://arxiv.org/abs/2608.24696}.

\bibitem[Lin et~al.(2025)Lin, Gou, Gong, Liu, Shen, Xu, Lin, Yang, Jiao, Duan,
  and Chen]{lin2024rho1}
Zhenghao Lin, Zhibin Gou, Yeyun Gong, Xiao Liu, Yelong Shen, Ruochen Xu, Chen
  Lin, Yujiu Yang, Jian Jiao, Nan Duan, and Weizhu Chen.
\newblock Rho-1: Not all tokens are what you need, 2025.
\newblock URL \url{https://arxiv.org/abs/2404.07965}.

\bibitem[Shao et~al.(2024)Shao, Wang, Zhu, Xu, Song, Bi, Zhang, Zhang, Li, Wu,
  and Guo]{shao2024deepseekmath}
Zhihong Shao, Peiyi Wang, Qihao Zhu, Runxin Xu, Junxiao Song, Xiao Bi, Haowei
  Zhang, Mingchuan Zhang, Y.~K. Li, Y.~Wu, and Daya Guo.
\newblock Deepseekmath: Pushing the limits of mathematical reasoning in open
  language models, 2024.
\newblock URL \url{https://arxiv.org/abs/2402.03300}.

\bibitem[Sheng et~al.(2026)Sheng, Ye, Wang, Wang, Gu, and Kang]{sheng2026ier}
Huanxin Sheng, Zhiling Ye, Haonan Wang, Jian Wang, Jinjie Gu, and Jian Kang.
\newblock 1
  distillation, 2026.
\newblock URL \url{https://arxiv.org/abs/2609.24432}.

\bibitem[Wang et~al.(2025{\natexlab{a}})Wang, Yang, Wang, Wang, Xu, and
  Huang]{wang2025abkd}
Guanghui Wang, Zhiyong Yang, Zitai Wang, Shi Wang, Qianqian Xu, and Qingming
  Huang.
\newblock Abkd: Pursuing a proper allocation of the probability mass in
  knowledge distillation via $\alpha$-$\beta$-divergence, 2025{\natexlab{a}}.
\newblock URL \url{https://arxiv.org/abs/2505.04560}.

\bibitem[Wang et~al.(2025{\natexlab{b}})Wang, Yu, Gao, Zheng, Liu, Lu, Dang,
  Chen, Yang, Zhang, Liu, Yang, Zhao, Yue, Song, Yu, Huang, and
  Lin]{wang2025beyond8020}
Shenzhi Wang, Le~Yu, Chang Gao, Chujie Zheng, Shixuan Liu, Rui Lu, Kai Dang,
  Xionghui Chen, Jianxin Yang, Zhenru Zhang, Yuqiong Liu, An~Yang, Andrew Zhao,
  Yang Yue, Shiji Song, Bowen Yu, Gao Huang, and Junyang Lin.
\newblock Beyond the 80/20 rule: High-entropy minority tokens drive effective
  reinforcement learning for llm reasoning, 2025{\natexlab{b}}.
\newblock URL \url{https://arxiv.org/abs/2506.01939}.

\bibitem[Wu et~al.(2024)Wu, Tao, Wang, Yang, Zhao, and Wong]{wu2025akl}
Taiqiang Wu, Chaofan Tao, Jiahao Wang, Runming Yang, Zhe Zhao, and Ngai Wong.
\newblock Rethinking kullback-leibler divergence in knowledge distillation for
  large language models, 2024.
\newblock URL \url{https://arxiv.org/abs/2404.02657}.

\bibitem[Xing et~al.(2026)Xing, Wang, Gao, Li, and Tang]{xing2026trust}
Xingrun Xing, Haoqing Wang, Boyan Gao, Ziheng Li, and Yehui Tang.
\newblock Trust region on-policy distillation, 2026.
\newblock URL \url{https://arxiv.org/abs/2606.01249}.

\bibitem[Xu et~al.(2026{\natexlab{a}})Xu, Sang, Zhou, He, and
  Wang]{xu2026paced}
Yuanda Xu, Hejian Sang, Zhengze Zhou, Ran He, and Zhipeng Wang.
\newblock Paced: Distillation and on-policy self-distillation at the frontier
  of student competence, 2026{\natexlab{a}}.
\newblock URL \url{https://arxiv.org/abs/2603.11178}.

\bibitem[Xu et~al.(2026{\natexlab{b}})Xu, Sang, Zhou, He, Wang, and
  Geramifard]{sang2026beyond}
Yuanda Xu, Hejian Sang, Zhengze Zhou, Ran He, Zhipeng Wang, and Alborz
  Geramifard.
\newblock Beyond grpo and on-policy distillation: An empirical sparse-to-dense
  reward principle for language-model post-training, 2026{\natexlab{b}}.
\newblock URL \url{https://arxiv.org/abs/2605.12483}.

\bibitem[Xu et~al.(2026{\natexlab{c}})Xu, Sang, Zhou, He, Wang, and
  Geramifard]{xu2026tip}
Yuanda Xu, Hejian Sang, Zhengze Zhou, Ran He, Zhipeng Wang, and Alborz
  Geramifard.
\newblock Tip: Token importance in on-policy distillation, 2026{\natexlab{c}}.
\newblock URL \url{https://arxiv.org/abs/2604.14084}.

\bibitem[Xu et~al.(2026{\natexlab{d}})Xu, Zhou, Behdin, Markovic-Voronov, Sang,
  Li, Zhu, Du, Rahmattalabi, He, Na, Wang, and Geramifard]{xu2026trek}
Yuanda Xu, Zhengze Zhou, Kayhan Behdin, Jelena Markovic-Voronov, Hejian Sang,
  Xiaomin Li, Wenhui Zhu, Xinchen Du, Aida Rahmattalabi, Ran He, Sen Na,
  Zhipeng Wang, and Alborz Geramifard.
\newblock Trek: Distill to explore, reinforce to refine, 2026{\natexlab{d}}.
\newblock URL \url{https://arxiv.org/abs/2607.05339}.

\bibitem[Yan et~al.(2025)Yan, Li, Hu, Wang, Cui, Qu, Cheng, and
  Zhang]{yan2025luffy}
Jianhao Yan, Yafu Li, Zican Hu, Zhi Wang, Ganqu Cui, Xiaoye Qu, Yu~Cheng, and
  Yue Zhang.
\newblock Learning to reason under off-policy guidance, 2025.
\newblock URL \url{https://arxiv.org/abs/2504.14945}.

\bibitem[Yang et~al.(2025)Yang, Li, Yang, Zhang, Hui, Zheng, Yu, Gao, Huang,
  Lv, Zheng, Liu, Zhou, Huang, Hu, Ge, Wei, Lin, Tang, Yang, Tu, Zhang, Yang,
  Yang, Zhou, Zhou, Lin, Dang, Bao, Yang, Yu, Deng, Li, Xue, Li, Zhang, Wang,
  Zhu, Men, Gao, Liu, Luo, Li, Tang, Yin, Ren, Wang, Zhang, Ren, Fan, Su,
  Zhang, Zhang, Wan, Liu, Wang, Cui, Zhang, Zhou, and Qiu]{yang2025qwen3}
An~Yang, Anfeng Li, Baosong Yang, Beichen Zhang, Binyuan Hui, Bo~Zheng, Bowen
  Yu, Chang Gao, Chengen Huang, Chenxu Lv, Chujie Zheng, Dayiheng Liu, Fan
  Zhou, Fei Huang, Feng Hu, Hao Ge, Haoran Wei, Huan Lin, Jialong Tang, Jian
  Yang, Jianhong Tu, Jianwei Zhang, Jianxin Yang, Jiaxi Yang, Jing Zhou,
  Jingren Zhou, Junyang Lin, Kai Dang, Keqin Bao, Kexin Yang, Le~Yu, Lianghao
  Deng, Mei Li, Mingfeng Xue, Mingze Li, Pei Zhang, Peng Wang, Qin Zhu, Rui
  Men, Ruize Gao, Shixuan Liu, Shuang Luo, Tianhao Li, Tianyi Tang, Wenbiao
  Yin, Xingzhang Ren, Xinyu Wang, Xinyu Zhang, Xuancheng Ren, Yang Fan, Yang
  Su, Yichang Zhang, Yinger Zhang, Yu~Wan, Yuqiong Liu, Zekun Wang, Zeyu Cui,
  Zhenru Zhang, Zhipeng Zhou, and Zihan Qiu.
\newblock Qwen3 technical report, 2025.
\newblock URL \url{https://arxiv.org/abs/2505.09388}.

\bibitem[Yue et~al.(2025)Yue, Chen, Lu, Zhao, Wang, Yue, Song, and
  Huang]{yue2025does}
Yang Yue, Zhiqi Chen, Rui Lu, Andrew Zhao, Zhaokai Wang, Yang Yue, Shiji Song,
  and Gao Huang.
\newblock Does reinforcement learning really incentivize reasoning capacity in
  llms beyond the base model?, 2025.
\newblock URL \url{https://arxiv.org/abs/2504.13837}.

\bibitem[Zhang et~al.(2025)Zhang, Li, Song, Ye, Gao, and Hu]{zhang2025law}
Chen Zhang, Qiuchi Li, Dawei Song, Zheyu Ye, Yan Gao, and Yan Hu.
\newblock Towards the law of capacity gap in distilling language models, 2025.
\newblock URL \url{https://arxiv.org/abs/2311.07052}.

\bibitem[Zhang et~al.(2026)Zhang, Sun, Zhao, Peng, Liang, Deng, Hu, Wong, and
  Chuan]{zhang2026verifydistill}
Zhiwei Zhang, Zechen Sun, Fei Zhao, Kang Peng, Bin Liang, Huayu Deng, Yao Hu,
  Kam-Fai Wong, and Mu~Chuan.
\newblock Verify before you distill: Prompt-level teacher gating for on-policy
  distillation, 2026.
\newblock URL \url{https://arxiv.org/abs/2609.02998}.

\end{thebibliography}
\bibliographystyle{iclr2027_conference}

\clearpage
\appendix

\section{Method Details}
\label{app:method}

\subsection{Calibrating the Entropy Brake}
\label{app:beta}

The coefficient $\beta$ in Eq.~\ref{eq:brake} is set by
\begin{equation}
\beta=0.1\cdot\frac{\mathrm{mean}_{\,m^B_t=1}\,|r_t|}{\mathrm{mean}_{\,m^B_t=1}\,H_S(t)}.
\label{eq:beta}
\end{equation}
We compute it once before training for each teacher--student pair and keep it
fixed. We take up to 50 student responses on C problems, use only the first
400 tokens of each, and skip responses shorter than 40 tokens. No new text is
generated. Both models score these responses in a forward pass: the teacher
gives its top 512 tokens plus the tail cell, and the student gives its
probabilities on the same 513 cells. From these we compute $H_S$, $H_T$,
$H_S-H_T$, and $|r_t|$ at every position. We then take the positions whose
$H_S-H_T$ is in the top 20\%, average $|r_t|$ and $H_S$ over them, and apply
Eq.~\ref{eq:beta}. Because $H_S$ is computed on $K_{\mathrm{tok}}+1$ cells, a
larger $K_{\mathrm{tok}}$ gives a larger $H_S$ and a smaller $\beta$; $\beta$
therefore depends on both the pair and $K_{\mathrm{tok}}$.
Table~\ref{tab:beta} lists the values used.

\begin{table}[h]
\centering
\small
\caption{Entropy-brake coefficient $\beta$ for each teacher--student pair.}
\label{tab:beta}
\begin{tabular}{lrr}
\toprule
Student & base-8B teacher & GRPO-8B teacher \\
\midrule
0.6B & 0.078 & 0.079 \\
1.7B & 0.058 & 0.063 \\
4B & 0.053 & 0.054 \\
\bottomrule
\end{tabular}
\end{table}

\subsection{Reverse KL from Sampled Tokens}
\label{app:rkl}

At one position, write $r(v)=\log\frac{\pi_S(v)}{\pi_T(v)}$, so
$D_{\mathrm R}=\sum_v\pi_S(v)\,r(v)$ and $r_t=r(y_t)$. Let $\theta$ be the
student parameters; the teacher is fixed. By the product rule,
\begin{equation}
\nabla_\theta D_{\mathrm R}=\sum_v\nabla_\theta\pi_S(v)\,r(v)+\sum_v\pi_S(v)\,\nabla_\theta r(v).
\end{equation}
The second term is zero. Because $\pi_T$ does not depend on $\theta$,
$\nabla_\theta r(v)=\nabla_\theta\log\pi_S(v)$, and
$\sum_v\pi_S(v)\nabla_\theta\log\pi_S(v)=\sum_v\nabla_\theta\pi_S(v)
=\nabla_\theta\sum_v\pi_S(v)=\nabla_\theta 1=0$. In the first term we use
$\nabla_\theta\pi_S=\pi_S\nabla_\theta\log\pi_S$, which gives
\begin{equation}
\nabla_\theta D_{\mathrm R}=\sum_v\pi_S(v)\,r(v)\,\nabla_\theta\log\pi_S(v)
=\mathbb E_{v\sim\pi_S}\bigl[r(v)\,\nabla_\theta\log\pi_S(v)\bigr].
\end{equation}
So for a sampled token $y_t$, the product $r_t\nabla_\theta\log\pi_S(y_t\mid s_t)$
is a one-sample estimate of $\nabla_\theta D_{\mathrm R}$. With $r_t$ held
constant, Eq.~\ref{eq:loss-c} has exactly this gradient at the selected
positions, so minimizing it lowers reverse KL there.

\subsection{Training Algorithm}

\begin{algorithm}[H]
\caption{\tsopd{} training}
\label{alg:tsopd}
\begin{algorithmic}[1]
\Require training set $\mathcal D_{\mathrm{train}}$, teacher $T_1$, raw student $S_0$
\State Sample $K_{\mathrm{cal}}$ responses per problem from $T_1$ and $S_0$; build $\mathcal D_A$, $\mathcal D_C$, and $\mathcal D_+=\mathcal D_A\cup\mathcal D_C$
\State Calibrate $\beta$ on student responses to C problems (Appendix~\ref{app:beta})
\For{each step}
  \State Draw a batch from $\mathcal D_+$; the student generates one response $y$ per problem
  \State The teacher scores $y$: top-$K_{\mathrm{tok}}$ probabilities and the tail cell at each position
  \If{$x\in\mathcal D_A$} compute $m^A_t$ and $\mathcal L^A$ (Eq.~\ref{eq:gate-a})
  \Else\ compute $m^C_t$, $\omega_t$, $m^B_t$, and $\mathcal L^C$ (Eqs.~\ref{eq:gate-c}--\ref{eq:brake})
  \EndIf
  \State Stop gradients through the gates; average the losses over the batch and make one update
\EndFor
\end{algorithmic}
\end{algorithm}

\section{Experimental Setup and Compute}
\label{app:setup}

\paragraph{Models and data.}
Students are Qwen3-0.6B, Qwen3-1.7B, and Qwen3-4B; the teachers are Qwen3-8B
(base-8B) and the same model after one epoch of GRPO (GRPO-8B). All training
runs use the MATH training set of 7{,}496 problems. Direct GRPO, OPD-r, and
OPD-f train on the full set, and their (A$\cup$C) variants train on the same
retained pool as \tsopd{}; \tsopd{} (routing only) and \tsopd{} train only on the
retained pool $\mathcal D_{\mathrm A}\cup\mathcal D_{\mathrm C}$ of the
corresponding teacher--student pair, screened with $K_{\mathrm{cal}}=8$
samples per model. Every run makes one pass over its training data, and we
evaluate the final checkpoint.

\paragraph{Framework.}
Training uses slime with a Megatron-LM training backend and SGLang for
rollout and teacher scoring. The chat template is applied with thinking mode
disabled for both training and evaluation.

\paragraph{Evaluation.}
MATH-test (5{,}000 problems) uses one sample per problem; MATH500, Minerva, and
OlympiadBench use eight; AIME24, AIME25, HMMT-Feb25, and HMMT-Nov25 use
sixteen. Sampling uses temperature 1.0 with a response limit of 8{,}192 tokens
for the first four benchmarks and 16{,}384 tokens for the competition sets.
Truncated or unparsable responses count as incorrect.

\paragraph{Compute.}
All training and evaluation runs in this paper used 1{,}728 NVIDIA B300 GPU
hours in total.

\section{Training Configurations}
\label{app:config}

\begin{table}[h]
\centering
\small
\caption{Training hyperparameters. Shared settings apply to all runs.}
\label{tab:app-config}
\begin{tabular}{lp{0.55\linewidth}}
\toprule
\multicolumn{2}{l}{\textit{Shared}} \\
Optimizer & Adam ($\beta_1=0.9$, $\beta_2=0.98$), weight decay 0.1 \\
Learning rate & $1\times10^{-6}$, constant \\
Responses per update & 64 \\
Rollout & temperature 1.0, max response length 8{,}192 \\
Policy clipping & $\epsilon_{\text{low}}=0.2$, $\epsilon_{\text{high}}=0.28$ \\
KL / entropy regularizer & none (coefficients 0) \\
\midrule
\multicolumn{2}{l}{\textit{Direct GRPO}} \\
Batch & 16 prompts $\times$ 4 responses \\
Reward & binary answer correctness \\
Steps & 469 (one epoch) \\
\midrule
\multicolumn{2}{l}{\textit{OPD-r and OPD-f}} \\
Batch & 64 prompts $\times$ 1 response \\
Objective & OPD-r: sampled-token reverse-KL penalty (coefficient 1.0);
OPD-f: forward KL on the teacher top-$K_{\mathrm{tok}}$ partition \\
Steps & 118 (one epoch) \\
\midrule
\multicolumn{2}{l}{\textit{\tsopd{} (routing only) and \tsopd{}}} \\
Batch & 64 prompts $\times$ 1 response \\
Training pool & $\mathcal D_{\mathrm A}\cup\mathcal D_{\mathrm C}$ of each pair, one epoch \\
Objective & forward KL on $A$, sampled reverse KL on $C$; \tsopd{} adds token
gates and the entropy brake \\
Token support & teacher top $K_{\mathrm{tok}}=512$ plus one tail cell \\
Gates (\tsopd{} only) & within-response quantile $0.8$; mass@$n>0.9$ with
$n=\max(\lfloor\ln K_{\mathrm{tok}}\rfloor,5)=6$; brake coefficient $\beta$
measured per pair (Eq.~\ref{eq:beta}) \\
\bottomrule
\end{tabular}
\end{table}

\clearpage
\section{Experimental Supplement}

\begin{table}[h]
\centering
\small
\caption{Training arms. All arms except the two Direct GRPO rows use the GRPO-8B
teacher; Table~\ref{tab:main} also reports the four OPD objectives with the
base-8B teacher.}
\label{tab:arms}
\begin{tabular}{lll}
\toprule
Arm & Training problems & Loss \\
\midrule
Direct GRPO & all & correctness reward, no teacher \\
Direct GRPO (\AC) & \AC & same \\
OPD-r & all & reverse KL \\
OPD-f & all & forward KL \\
OPD-r (\AC) & \AC & reverse KL \\
OPD-f (\AC) & \AC & forward KL \\
\tsro{} & \AC & forward KL on A, reverse KL on C; no gates, no brake \\
\tsopd{} & \AC & same, plus token gates and the entropy brake \\
\bottomrule
\end{tabular}
\end{table}

\begin{table}[h]
\centering
\small
\caption{Ablations with the 0.6B student and the GRPO-8B teacher (macro averages
over eight benchmarks, \%). Each row changes one setting of the default
\tsopd{}.}
\label{tab:ablation}
\begin{tabular}{lrr}
\toprule
Setting & \avgk{} & \passk{} \\
\midrule
$\tau=0.60$ & 27.17 & 41.45 \\
$\tau=0.95$ & 24.71 & 36.41 \\
20\% of positions by student entropy & 26.28 & 40.54 \\
20\% of positions at random & 26.36 & 41.34 \\
$\mathrm{mass@}n>0.0$ & 26.70 & 39.93 \\
$\mathrm{mass@}n>0.5$ & 26.75 & 42.59 \\
$\mathrm{mass@}n>0.7$ & 26.66 & 41.11 \\
$n=1$ & 25.76 & 38.26 \\
$n=3$ & 27.27 & 42.70 \\
$n=10$ & 25.82 & 38.53 \\
$\beta=0$ & 26.87 & 41.38 \\
\midrule
\tsopd{} (default) & \textbf{28.88} & \textbf{44.57} \\
\bottomrule
\end{tabular}
\end{table}

\begin{figure}[h]
\centering
\includegraphics[width=\linewidth]{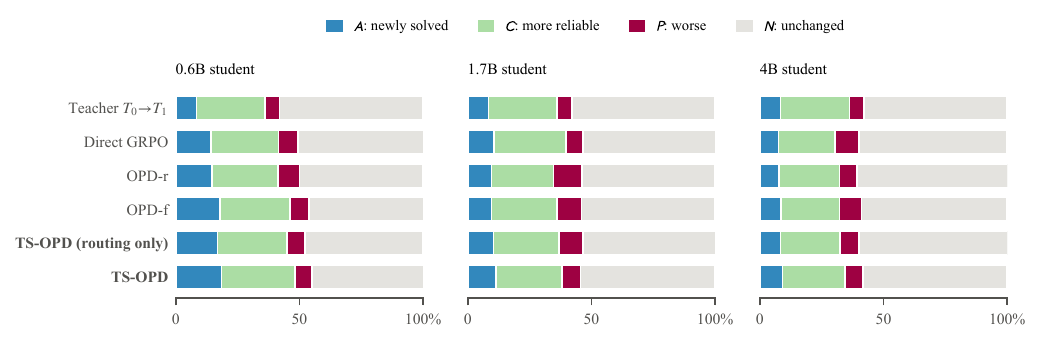}
\caption{Outcome categories of all 1{,}566 test problems after training.
Students are compared with $S_0$; the first row shows the teacher from $T_0$ to
$T_1$.}
\label{fig:acpn-all}
\end{figure}

\begin{table}[h]
\centering
\scriptsize
\setlength{\tabcolsep}{3pt}
\caption{Outcome categories of all 1{,}566 benchmark problems after training (\%), relative to the raw student $S_0$; the teacher row is $T_0\to T_1$.}
\label{tab:app-acpn}
\begin{tabular}{lrrrrrrrrrrrr}
\toprule
 & \multicolumn{4}{c}{0.6B} & \multicolumn{4}{c}{1.7B} & \multicolumn{4}{c}{4B} \\
\cmidrule(lr){2-5}\cmidrule(lr){6-9}\cmidrule(lr){10-13}
Method & A & C & P & N & A & C & P & N & A & C & P & N \\
\midrule
Teacher $T_0\!\to\!T_1$ & 8.4 & 27.8 & 5.8 & 58.0 & 8.4 & 27.8 & 5.8 & 58.0 & 8.4 & 27.8 & 5.8 & 58.0 \\
Direct GRPO & 14.3 & 27.3 & 7.9 & 50.6 & 10.7 & 28.9 & 6.9 & 53.6 & 7.5 & 22.8 & 9.8 & 59.8 \\
Direct GRPO (A$\cup$C) & 13.5 & 27.0 & 8.0 & 51.5 & 10.2 & 28.3 & 7.2 & 54.3 & 7.0 & 22.3 & 9.8 & 60.9 \\
OPD-r & 14.6 & 26.8 & 8.6 & 50.0 & 9.6 & 25.0 & 11.7 & 53.7 & 7.9 & 24.3 & 7.2 & 60.7 \\
OPD-r (A$\cup$C) & 15.8 & 28.7 & 7.3 & 48.1 & 10.0 & 27.3 & 9.1 & 53.6 & 7.0 & 24.1 & 7.9 & 61.0 \\
OPD-f & 17.9 & 28.3 & 7.7 & 46.2 & 9.6 & 26.5 & 9.9 & 54.0 & 8.7 & 23.5 & 8.9 & 58.9 \\
OPD-f (A$\cup$C) & 14.2 & 27.2 & 9.3 & 49.3 & 9.1 & 25.4 & 10.9 & 54.6 & 7.2 & 23.1 & 9.0 & 60.7 \\
TS-OPD (routing only) & 16.9 & 28.1 & 7.2 & 47.8 & 10.4 & 26.4 & 9.9 & 53.4 & 8.4 & 24.1 & 7.6 & 60.0 \\
TS-OPD & 18.5 & 29.8 & 6.9 & 44.9 & 11.3 & 26.9 & 7.3 & 54.5 & 9.2 & 25.2 & 7.2 & 58.4 \\
\bottomrule
\end{tabular}
\end{table}

\begin{table}[h]
\centering
\scriptsize
\setlength{\tabcolsep}{3pt}
\caption{Outcome categories, in the student, of the 567 problems that GRPO improves for the teacher (\%).}
\label{tab:app-rq3}
\begin{tabular}{lrrrrrrrrrrrr}
\toprule
 & \multicolumn{4}{c}{0.6B} & \multicolumn{4}{c}{1.7B} & \multicolumn{4}{c}{4B} \\
\cmidrule(lr){2-5}\cmidrule(lr){6-9}\cmidrule(lr){10-13}
Method & A & C & P & N & A & C & P & N & A & C & P & N \\
\midrule
Direct GRPO & 24.3 & 16.9 & 8.8 & 49.9 & 23.3 & 41.4 & 6.7 & 28.6 & 17.3 & 45.1 & 10.8 & 26.8 \\
Direct GRPO (A$\cup$C) & 21.0 & 16.1 & 9.0 & 54.0 & 20.5 & 39.0 & 7.6 & 33.0 & 14.5 & 43.6 & 11.0 & 31.0 \\
OPD-r & 25.0 & 16.2 & 7.9 & 50.8 & 21.7 & 38.9 & 8.6 & 30.8 & 18.7 & 47.6 & 7.4 & 26.3 \\
OPD-r (A$\cup$C) & 28.0 & 18.0 & 7.9 & 46.0 & 23.3 & 39.9 & 8.6 & 28.2 & 15.9 & 47.4 & 8.5 & 28.2 \\
OPD-f & 29.1 & 20.3 & 6.5 & 44.1 & 22.4 & 35.6 & 11.3 & 30.7 & 18.2 & 46.9 & 8.8 & 26.1 \\
OPD-f (A$\cup$C) & 24.2 & 17.1 & 9.5 & 49.2 & 21.7 & 36.2 & 11.1 & 31.0 & 14.8 & 46.2 & 10.4 & 28.6 \\
TS-OPD (routing only) & 30.7 & 18.0 & 7.6 & 43.7 & 23.9 & 38.1 & 9.3 & 28.7 & 18.3 & 49.0 & 7.6 & 25.0 \\
TS-OPD & 32.1 & 18.0 & 8.6 & 41.3 & 24.0 & 38.1 & 8.2 & 29.8 & 19.2 & 49.2 & 7.2 & 24.3 \\
\bottomrule
\end{tabular}
\end{table}

\begin{table}[h]
\centering
\scriptsize
\setlength{\tabcolsep}{2.8pt}
\caption{Outcome categories (\%) of test problems where the teacher is not
better (Figure~\ref{fig:where}(c, d)). P stratum: $S_0$ more accurate than $T_1$ (41,
65, and 103 problems for 0.6B, 1.7B, and 4B). N stratum: equally accurate (453,
711, and 909 problems). Each row sums to 100 within a size; ``-'' marks a
category that cannot occur. All arms except the two Direct GRPO rows use the GRPO-8B
teacher.}
\label{tab:rq2}
\begin{tabular}{lrrrrrrrrrrrr}
\toprule
 & \multicolumn{4}{c}{0.6B} & \multicolumn{4}{c}{1.7B} & \multicolumn{4}{c}{4B} \\
\cmidrule(lr){2-5}\cmidrule(lr){6-9}\cmidrule(lr){10-13}
Method & A & C & P & N & A & C & P & N & A & C & P & N \\
\midrule
\multicolumn{13}{l}{\textit{P stratum: $S_0$ more accurate than $T_1$}} \\
Direct GRPO & - & 12.2 & 65.9 & 22.0 & - & 20.0 & 49.2 & 30.8 & - & 6.8 & 72.8 & 20.4 \\
Direct GRPO (A$\cup$C) & - & 14.6 & 63.4 & 22.0 & - & 18.5 & 50.8 & 30.8 & - & 7.8 & 71.8 & 20.4 \\
OPD-r & - & 14.6 & 78.0 & 7.3 & - & 16.9 & 70.8 & 12.3 & - & 13.6 & 58.3 & 28.2 \\
OPD-r (A$\cup$C) & - & 17.1 & 73.2 & 9.8 & - & 13.8 & 64.6 & 21.5 & - & 12.6 & 68.9 & 18.4 \\
OPD-f & - & 4.9 & 82.9 & 12.2 & - & 15.4 & 63.1 & 21.5 & - & 15.5 & 55.3 & 29.1 \\
OPD-f (A$\cup$C) & - & 7.3 & 78.0 & 14.6 & - & 16.9 & 66.2 & 16.9 & - & 7.8 & 65.0 & 27.2 \\
TS-OPD (routing only) & - & 17.1 & 65.9 & 17.1 & - & 20.0 & 60.0 & 20.0 & - & 7.8 & 60.2 & 32.0 \\
TS-OPD & - & 14.6 & 73.2 & 12.2 & - & 13.8 & 70.8 & 15.4 & - & 9.7 & 66.0 & 24.3 \\
\midrule
\multicolumn{13}{l}{\textit{N stratum: $S_0$ and $T_1$ equally accurate}} \\
Direct GRPO & 2.6 & 0.4 & 2.4 & 94.5 & 3.1 & 1.4 & 3.5 & 92.0 & 1.7 & 1.4 & 5.4 & 91.5 \\
Direct GRPO (A$\cup$C) & 2.4 & 0.4 & 2.4 & 94.7 & 3.0 & 1.3 & 3.7 & 92.1 & 1.5 & 1.3 & 5.3 & 91.9 \\
OPD-r & 4.0 & 0.2 & 3.1 & 92.7 & 2.4 & 0.7 & 6.0 & 90.8 & 1.5 & 2.2 & 3.7 & 92.5 \\
OPD-r (A$\cup$C) & 2.0 & 0.2 & 2.9 & 94.9 & 1.7 & 0.6 & 5.9 & 91.8 & 1.8 & 1.9 & 3.2 & 93.2 \\
OPD-f & 3.1 & 0.4 & 4.9 & 91.6 & 1.5 & 0.4 & 8.3 & 89.7 & 3.1 & 2.1 & 5.4 & 89.4 \\
OPD-f (A$\cup$C) & 3.3 & 0.2 & 4.2 & 92.3 & 1.4 & 0.6 & 7.0 & 91.0 & 2.5 & 1.5 & 5.3 & 90.6 \\
TS-OPD (routing only) & 3.8 & 0.4 & 2.4 & 93.4 & 1.9 & 0.6 & 5.6 & 91.9 & 2.4 & 1.9 & 4.2 & 91.5 \\
TS-OPD & 5.1 & 0.7 & 2.4 & 91.8 & 2.2 & 0.7 & 5.1 & 92.0 & 2.3 & 2.5 & 3.7 & 91.4 \\
\bottomrule
\end{tabular}

\end{table}

\begin{table}[h]
\centering
\scriptsize
\setlength{\tabcolsep}{3.5pt}
\caption{Outcome categories (\%) of knowledge-gap and precision-gap test
problems (Figure~\ref{fig:where}(b)). All arms except the two Direct GRPO rows use the
GRPO-8B teacher; the \AC{} rows share the same problems and steps. Knowledge-gap problems start at 0, so they can
only be $\tA$ or $\tN$; precision-gap problems start above 0, so they can only
be $\tC$, $\tP$, or $\tN$. There are 517 and 555 such problems for 0.6B, 264 and
526 for 1.7B, and 152 and 402 for 4B.}
\label{tab:rq4}
\begin{tabular}{lrrrrrrrrrrrr}
\toprule
 & \multicolumn{4}{c}{0.6B} & \multicolumn{4}{c}{1.7B} & \multicolumn{4}{c}{4B} \\
\cmidrule(lr){2-5}\cmidrule(lr){6-9}\cmidrule(lr){10-13}
Method & A & C & P & N & A & C & P & N & A & C & P & N \\
\midrule
\multicolumn{13}{l}{\textit{Knowledge-gap problems: $S_0$ at 0, $T_1$ above 0}} \\
Direct GRPO & 41.0 & - & - & 59.0 & 54.9 & - & - & 45.1 & 67.8 & - & - & 32.2 \\
Direct GRPO (A$\cup$C) & 38.1 & - & - & 61.9 & 51.9 & - & - & 48.1 & 59.9 & - & - & 40.1 \\
OPD-r & 41.6 & - & - & 58.4 & 52.7 & - & - & 47.3 & 71.1 & - & - & 28.9 \\
OPD-r (A$\cup$C) & 40.2 & - & - & 59.8 & 50.4 & - & - & 49.6 & 59.2 & - & - & 40.8 \\
OPD-f & 50.7 & - & - & 49.3 & 50.3 & - & - & 49.7 & 71.7 & - & - & 28.3 \\
OPD-f (A$\cup$C) & 46.2 & - & - & 53.8 & 54.5 & - & - & 45.5 & 61.8 & - & - & 38.2 \\
TS-OPD (routing only) & 47.0 & - & - & 53.0 & 56.2 & - & - & 43.8 & 71.7 & - & - & 28.3 \\
TS-OPD & 48.0 & - & - & 52.0 & 55.3 & - & - & 44.7 & 75.7 & - & - & 24.3 \\
\midrule
\multicolumn{13}{l}{\textit{Precision-gap problems: $0<S_0<T_1$}} \\
Direct GRPO & - & 75.7 & 15.3 & 9.0 & - & 81.6 & 9.7 & 8.7 & - & 83.8 & 7.5 & 8.7 \\
Direct GRPO (A$\cup$C) & - & 73.5 & 17.1 & 9.4 & - & 78.5 & 11.4 & 10.1 & - & 81.6 & 9.0 & 9.5 \\
OPD-r & - & 78.6 & 13.3 & 8.1 & - & 76.0 & 12.5 & 11.5 & - & 86.1 & 4.5 & 9.5 \\
OPD-r (A$\cup$C) & - & 79.5 & 13.0 & 7.6 & - & 78.9 & 11.0 & 10.1 & - & 86.6 & 5.7 & 7.7 \\
OPD-f & - & 74.2 & 14.8 & 11.0 & - & 72.2 & 14.8 & 12.9 & - & 84.3 & 5.5 & 10.2 \\
OPD-f (A$\cup$C) & - & 76.0 & 16.9 & 7.0 & - & 72.6 & 14.8 & 12.5 & - & 84.6 & 6.5 & 9.0 \\
TS-OPD (routing only) & - & 77.7 & 13.3 & 9.0 & - & 75.3 & 14.4 & 10.4 & - & 87.6 & 4.7 & 7.7 \\
TS-OPD & - & 83.1 & 11.4 & 5.6 & - & 77.3 & 10.0 & 12.7 & - & 88.6 & 5.5 & 6.0 \\
\bottomrule
\end{tabular}

\end{table}
\begin{table}[h]
\centering
\scriptsize
\setlength{\tabcolsep}{3pt}
\caption{Transfer gap $G_{\mathrm{transfer}}$ (percentage points) on each benchmark with $K\ge8$. The last row per student is the macro average used in Table~\ref{tab:main}.}
\label{tab:app-gap}
\setlength{\aboverulesep}{0pt}\setlength{\belowrulesep}{0pt}\renewcommand{\arraystretch}{1.12}
\begin{tabular}{llrrrrrrrr|r}
\toprule
 & & \multicolumn{4}{c}{base-8B teacher} & \multicolumn{4}{c|}{GRPO-8B teacher} & No teacher \\
\cmidrule(lr){3-6}\cmidrule(lr){7-10}
Student & Benchmark & OPD-r & OPD-f & \shortstack{TS-OPD\\(routing\\only)} & TS-OPD & OPD-r & OPD-f & \shortstack{TS-OPD\\(routing\\only)} & TS-OPD & \shortstack{Direct\\GRPO} \\
\midrule
0.6B & MATH500 & -3.05 & -1.45 & -5.92 & -6.64 & -14.42 & -12.17 & -15.22 & -18.07 & -12.55 \\
 & Minerva & 6.20 & 5.87 & 5.32 & 4.31 & -1.52 & -0.74 & -0.47 & -0.79 & -0.38 \\
 & OlympiadBench & 4.63 & 5.58 & 3.11 & 2.26 & -4.62 & -0.99 & -4.14 & -4.94 & -1.12 \\
 & AIME24 & 30.62 & 28.75 & 29.58 & 27.08 & 24.16 & 23.54 & 21.46 & 22.50 & 24.58 \\
 & AIME25 & 26.05 & 26.46 & 23.96 & 22.92 & 15.00 & 14.17 & 13.55 & 12.09 & 14.79 \\
 & HMMT-Feb25 & 15.21 & 15.84 & 14.59 & 14.59 & 8.55 & 10.00 & 8.75 & 5.42 & 10.42 \\
 & HMMT-Nov25 & 26.05 & 27.51 & 26.26 & 26.89 & 24.59 & 23.76 & 24.59 & 23.55 & 25.22 \\
\cmidrule(lr){2-11}
 & Macro & 15.10 & 15.51 & 13.84 & 13.06 & 7.39 & 8.22 & 6.93 & 5.68 & 8.71 \\
\midrule
1.7B & MATH500 & 6.42 & 9.63 & 6.63 & 5.70 & -2.12 & -1.00 & -1.95 & -2.45 & -3.58 \\
 & Minerva & 8.45 & 10.52 & 7.81 & 6.80 & 1.47 & 0.96 & 1.14 & 0.50 & -0.19 \\
 & OlympiadBench & 11.09 & 13.52 & 10.18 & 10.07 & 1.57 & 3.04 & 1.22 & 0.87 & -1.65 \\
 & AIME24 & 27.08 & 30.83 & 30.20 & 27.91 & 12.49 & 12.70 & 9.58 & 9.16 & 12.08 \\
 & AIME25 & 24.17 & 24.59 & 24.80 & 24.17 & 11.47 & 9.17 & 11.88 & 10.21 & 11.05 \\
 & HMMT-Feb25 & 16.88 & 16.88 & 14.59 & 15.21 & 6.67 & 6.25 & 5.63 & 6.04 & 8.75 \\
 & HMMT-Nov25 & 27.71 & 28.75 & 26.67 & 27.09 & 19.38 & 18.96 & 17.50 & 16.05 & 22.09 \\
\cmidrule(lr){2-11}
 & Macro & 17.40 & 19.25 & 17.27 & 16.71 & 7.28 & 7.15 & 6.43 & 5.77 & 6.94 \\
\midrule
4B & MATH500 & 9.13 & 10.30 & 9.10 & 8.37 & 0.28 & 0.78 & -0.15 & -0.78 & 0.78 \\
 & Minerva & 7.99 & 7.58 & 7.76 & 7.90 & 0.64 & 1.92 & 1.19 & -1.89 & 1.51 \\
 & OlympiadBench & 15.15 & 16.41 & 14.24 & 13.43 & 3.47 & 3.65 & 3.37 & 3.17 & 5.21 \\
 & AIME24 & 34.58 & 33.75 & 32.91 & 32.29 & 5.62 & 2.91 & 6.04 & 4.99 & 5.83 \\
 & AIME25 & 27.08 & 27.29 & 27.50 & 27.71 & 3.75 & 1.88 & 5.21 & -1.25 & 5.62 \\
 & HMMT-Feb25 & 16.67 & 16.88 & 16.04 & 15.21 & 0.42 & -0.62 & -3.33 & -1.25 & 0.83 \\
 & HMMT-Nov25 & 29.80 & 27.71 & 29.17 & 29.59 & 4.38 & 3.34 & 2.30 & 0.21 & 4.38 \\
\cmidrule(lr){2-11}
 & Macro & 20.06 & 19.99 & 19.53 & 19.21 & 2.65 & 1.98 & 2.09 & 0.46 & 3.45 \\
\bottomrule
\end{tabular}
\end{table}

\begin{table}[h]
\centering
\scriptsize
\setlength{\tabcolsep}{3.5pt}
\caption{Per-benchmark results for the 0.6B student (\%). Direct GRPO uses no teacher. Macro averages weight each benchmark equally.}
\label{tab:app-full-06B}
\begin{tabular}{llrrrrrrrrr}
\toprule
 & & & \multicolumn{4}{c}{Teacher: base-8B} & \multicolumn{4}{c}{Teacher: GRPO-8B} \\
\cmidrule(lr){4-7}\cmidrule(lr){8-11}
Benchmark & Metric & \shortstack{Direct\\GRPO} & OPD-r & OPD-f & \shortstack{TS-OPD\\(routing only)} & TS-OPD & OPD-r & OPD-f & \shortstack{TS-OPD\\(routing only)} & TS-OPD \\
\midrule
MATH500 & Avg@8 & 62.58 & 53.08 & 51.48 & 55.95 & 56.67 & 64.45 & 62.20 & 65.25 & 68.10 \\
 & Pass@8 & 82.00 & 78.80 & 78.00 & 82.80 & 81.60 & 85.00 & 82.60 & 84.40 & 85.40 \\
Minerva & Avg@8 & 16.87 & 10.29 & 10.62 & 11.17 & 12.18 & 18.01 & 17.23 & 16.96 & 17.28 \\
 & Pass@8 & 30.88 & 25.74 & 27.21 & 28.31 & 26.84 & 33.46 & 34.93 & 32.72 & 32.35 \\
OlympiadBench & Avg@8 & 28.86 & 23.11 & 22.16 & 24.63 & 25.48 & 32.36 & 28.73 & 31.88 & 32.68 \\
 & Pass@8 & 47.03 & 44.07 & 45.10 & 45.85 & 48.66 & 52.23 & 46.88 & 50.89 & 50.15 \\
AIME24 & Avg@16 & 7.50 & 1.46 & 3.33 & 2.50 & 5.00 & 7.92 & 8.54 & 10.62 & 9.58 \\
 & Pass@16 & 30.00 & 10.00 & 20.00 & 13.33 & 20.00 & 36.67 & 33.33 & 43.33 & 30.00 \\
AIME25 & Avg@16 & 14.38 & 3.12 & 2.71 & 5.21 & 6.25 & 14.17 & 15.00 & 15.62 & 17.08 \\
 & Pass@16 & 33.33 & 20.00 & 13.33 & 20.00 & 23.33 & 36.67 & 30.00 & 40.00 & 33.33 \\
HMMT-Feb25 & Avg@16 & 6.25 & 1.46 & 0.83 & 2.08 & 2.08 & 8.12 & 6.67 & 7.92 & 11.25 \\
 & Pass@16 & 20.00 & 10.00 & 10.00 & 10.00 & 16.67 & 20.00 & 16.67 & 20.00 & 26.67 \\
HMMT-Nov25 & Avg@16 & 4.79 & 3.96 & 2.50 & 3.75 & 3.12 & 5.42 & 6.25 & 5.42 & 6.46 \\
 & Pass@16 & 13.33 & 13.33 & 10.00 & 13.33 & 13.33 & 23.33 & 23.33 & 20.00 & 30.00 \\
MATH-test & Avg@1 & 67.10 & 60.42 & 59.60 & 59.52 & 59.34 & 65.54 & 67.92 & 67.30 & 68.62 \\
\midrule
Macro Avg & Avg@k & 26.04 & 19.61 & 19.15 & 20.60 & 21.27 & 27.00 & 26.57 & 27.62 & 28.88 \\
 & Pass@k & 40.46 & 32.80 & 32.91 & 34.14 & 36.22 & 44.11 & 41.96 & 44.83 & 44.57 \\
\bottomrule
\end{tabular}
\end{table}

\begin{table}[h]
\centering
\scriptsize
\setlength{\tabcolsep}{3.5pt}
\caption{Per-benchmark results for the 1.7B student (\%). Direct GRPO uses no teacher. Macro averages weight each benchmark equally.}
\label{tab:app-full-17B}
\begin{tabular}{llrrrrrrrrr}
\toprule
 & & & \multicolumn{4}{c}{Teacher: base-8B} & \multicolumn{4}{c}{Teacher: GRPO-8B} \\
\cmidrule(lr){4-7}\cmidrule(lr){8-11}
Benchmark & Metric & \shortstack{Direct\\GRPO} & OPD-r & OPD-f & \shortstack{TS-OPD\\(routing only)} & TS-OPD & OPD-r & OPD-f & \shortstack{TS-OPD\\(routing only)} & TS-OPD \\
\midrule
MATH500 & Avg@8 & 81.38 & 71.38 & 68.17 & 71.17 & 72.10 & 79.92 & 78.80 & 79.75 & 80.25 \\
 & Pass@8 & 93.00 & 90.40 & 89.60 & 90.40 & 89.80 & 91.80 & 91.20 & 92.00 & 92.00 \\
Minerva & Avg@8 & 29.00 & 20.36 & 18.29 & 21.00 & 22.01 & 27.34 & 27.85 & 27.67 & 28.31 \\
 & Pass@8 & 44.49 & 37.50 & 35.66 & 37.50 & 37.13 & 42.28 & 42.65 & 44.12 & 44.12 \\
OlympiadBench & Avg@8 & 48.94 & 36.20 & 33.77 & 37.11 & 37.22 & 45.72 & 44.25 & 46.07 & 46.42 \\
 & Pass@8 & 65.28 & 59.20 & 57.57 & 60.39 & 60.09 & 62.91 & 61.57 & 63.80 & 63.35 \\
AIME24 & Avg@16 & 29.58 & 14.58 & 10.83 & 11.46 & 13.75 & 29.17 & 28.96 & 32.08 & 32.50 \\
 & Pass@16 & 66.67 & 33.33 & 50.00 & 50.00 & 40.00 & 60.00 & 66.67 & 66.67 & 63.33 \\
AIME25 & Avg@16 & 23.54 & 10.42 & 10.00 & 9.79 & 10.42 & 23.12 & 25.42 & 22.71 & 24.38 \\
 & Pass@16 & 53.33 & 33.33 & 26.67 & 30.00 & 33.33 & 43.33 & 46.67 & 53.33 & 50.00 \\
HMMT-Feb25 & Avg@16 & 11.67 & 3.54 & 3.54 & 5.83 & 5.21 & 13.75 & 14.17 & 14.79 & 14.38 \\
 & Pass@16 & 26.67 & 10.00 & 16.67 & 23.33 & 16.67 & 33.33 & 33.33 & 30.00 & 40.00 \\
HMMT-Nov25 & Avg@16 & 9.58 & 3.96 & 2.92 & 5.00 & 4.58 & 12.29 & 12.71 & 14.17 & 15.62 \\
 & Pass@16 & 26.67 & 16.67 & 10.00 & 23.33 & 16.67 & 40.00 & 36.67 & 43.33 & 43.33 \\
MATH-test & Avg@1 & 82.54 & 73.28 & 73.04 & 72.94 & 72.40 & 81.00 & 81.82 & 81.48 & 82.12 \\
\midrule
Macro Avg & Avg@k & 39.53 & 29.22 & 27.57 & 29.29 & 29.71 & 39.04 & 39.25 & 39.84 & 40.50 \\
 & Pass@k & 57.33 & 44.21 & 44.90 & 48.49 & 45.76 & 56.83 & 57.57 & 59.34 & 59.78 \\
\bottomrule
\end{tabular}
\end{table}

\begin{table}[h]
\centering
\scriptsize
\setlength{\tabcolsep}{3.5pt}
\caption{Per-benchmark results for the 4B student (\%). Direct GRPO uses no teacher. Macro averages weight each benchmark equally.}
\label{tab:app-full-4B}
\begin{tabular}{llrrrrrrrrr}
\toprule
 & & & \multicolumn{4}{c}{Teacher: base-8B} & \multicolumn{4}{c}{Teacher: GRPO-8B} \\
\cmidrule(lr){4-7}\cmidrule(lr){8-11}
Benchmark & Metric & \shortstack{Direct\\GRPO} & OPD-r & OPD-f & \shortstack{TS-OPD\\(routing only)} & TS-OPD & OPD-r & OPD-f & \shortstack{TS-OPD\\(routing only)} & TS-OPD \\
\midrule
MATH500 & Avg@8 & 88.47 & 80.12 & 78.95 & 80.15 & 80.88 & 88.97 & 88.47 & 89.40 & 90.03 \\
 & Pass@8 & 94.80 & 93.60 & 93.40 & 93.40 & 93.20 & 95.40 & 94.60 & 95.60 & 95.20 \\
Minerva & Avg@8 & 35.80 & 29.32 & 29.73 & 29.55 & 29.41 & 36.67 & 35.39 & 36.12 & 39.20 \\
 & Pass@8 & 48.53 & 43.75 & 44.49 & 43.38 & 41.54 & 48.53 & 51.84 & 47.79 & 52.21 \\
OlympiadBench & Avg@8 & 55.10 & 45.16 & 43.90 & 46.07 & 46.88 & 56.84 & 56.66 & 56.94 & 57.14 \\
 & Pass@8 & 67.66 & 65.28 & 65.88 & 66.91 & 66.47 & 68.55 & 70.03 & 69.58 & 70.33 \\
AIME24 & Avg@16 & 48.54 & 19.79 & 20.62 & 21.46 & 22.08 & 48.75 & 51.46 & 48.33 & 49.38 \\
 & Pass@16 & 76.67 & 46.67 & 50.00 & 43.33 & 53.33 & 80.00 & 83.33 & 83.33 & 73.33 \\
AIME25 & Avg@16 & 39.38 & 17.92 & 17.71 & 17.50 & 17.29 & 41.25 & 43.12 & 39.79 & 46.25 \\
 & Pass@16 & 73.33 & 50.00 & 43.33 & 46.67 & 43.33 & 73.33 & 73.33 & 73.33 & 73.33 \\
HMMT-Feb25 & Avg@16 & 24.17 & 8.33 & 8.12 & 8.96 & 9.79 & 24.58 & 25.62 & 28.33 & 26.25 \\
 & Pass@16 & 40.00 & 20.00 & 23.33 & 23.33 & 20.00 & 46.67 & 56.67 & 66.67 & 53.33 \\
HMMT-Nov25 & Avg@16 & 31.25 & 5.83 & 7.92 & 6.46 & 6.04 & 31.25 & 32.29 & 33.33 & 35.42 \\
 & Pass@16 & 56.67 & 30.00 & 30.00 & 23.33 & 26.67 & 63.33 & 60.00 & 63.33 & 60.00 \\
MATH-test & Avg@1 & 89.86 & 82.12 & 82.04 & 82.10 & 81.96 & 89.86 & 89.64 & 89.84 & 90.56 \\
\midrule
Macro Avg & Avg@k & 51.57 & 36.07 & 36.12 & 36.53 & 36.79 & 52.27 & 52.83 & 52.76 & 54.28 \\
 & Pass@k & 68.44 & 53.93 & 54.06 & 52.81 & 53.31 & 70.71 & 72.43 & 73.68 & 71.04 \\
\bottomrule
\end{tabular}
\end{table}

\begin{table}[h]
\centering
\scriptsize
\caption{Per-benchmark results of the raw students and the two teachers (\%).}
\label{tab:app-base}
\begin{tabular}{llrrrrr}
\toprule
Benchmark & Metric & Qwen3-0.6B & Qwen3-1.7B & Qwen3-4B & base-8B & GRPO-8B \\
\midrule
MATH500 & Avg@8 & 42.23 & 70.00 & 81.45 & 82.30 & 90.10 \\
 & Pass@8 & 72.60 & 89.40 & 93.40 & 93.60 & 95.40 \\
Minerva & Avg@8 & 8.64 & 20.96 & 29.46 & 30.06 & 37.91 \\
 & Pass@8 & 25.74 & 35.66 & 41.54 & 42.65 & 50.00 \\
OlympiadBench & Avg@8 & 15.37 & 34.92 & 47.94 & 47.59 & 59.96 \\
 & Pass@8 & 37.09 & 56.82 & 66.02 & 67.21 & 72.26 \\
AIME24 & Avg@16 & 1.04 & 10.62 & 23.33 & 25.42 & 56.46 \\
 & Pass@16 & 10.00 & 30.00 & 46.67 & 60.00 & 86.67 \\
AIME25 & Avg@16 & 1.25 & 6.67 & 17.08 & 17.29 & 45.21 \\
 & Pass@16 & 13.33 & 23.33 & 40.00 & 36.67 & 73.33 \\
HMMT-Feb25 & Avg@16 & 0.21 & 3.96 & 8.54 & 9.79 & 26.25 \\
 & Pass@16 & 3.33 & 13.33 & 16.67 & 20.00 & 50.00 \\
HMMT-Nov25 & Avg@16 & 1.67 & 3.33 & 7.29 & 8.33 & 36.67 \\
 & Pass@16 & 10.00 & 13.33 & 26.67 & 30.00 & 63.33 \\
MATH-test & Avg@1 & 49.70 & 71.16 & 82.68 & 83.46 & 91.48 \\
\midrule
Macro Avg & Avg@k & 15.01 & 27.70 & 37.22 & 38.03 & 55.51 \\
 & Pass@k & 27.72 & 41.63 & 51.71 & 54.20 & 72.81 \\
\bottomrule
\end{tabular}
\end{table}

\end{document}